\documentclass[twoside]{article}

\usepackage[accepted]{aistats2023}
\usepackage[round]{natbib}

\usepackage[utf8]{inputenc} % allow utf-8 input
\usepackage[T1]{fontenc}    % use 8-bit T1 fonts
\usepackage[hidelinks]{hyperref}       % hyperlinks
\usepackage{url}            % simple URL typesetting
\usepackage{booktabs}       % professional-quality tables
\usepackage{amsfonts}       % blackboard math symbols
\usepackage{nicefrac}       % compact symbols for 1/2, etc.
\usepackage{microtype}      % microtypography
\usepackage{xcolor}         % colors
\usepackage{graphicx}
\usepackage{subfigure}

\usepackage{amssymb,amsmath}
\usepackage{amsthm}

\newcommand{\result}[2]{\makebox[2.5em]{\hfill#1}\makebox[2.7em]{\hfill(#2)}}
\newcommand{\resultbold}[2]{\makebox[2.5em]{\hfill\textbf{#1}}\makebox[2.7em]{\hfill\textbf{(#2)}}}
\newcommand{\resultblueit}[2]{\makebox[2.5em]{\hfill\emph{\color{blue} #1}}\makebox[2.7em]{\hfill\emph{\color{blue} (#2)}}}

\newcommand{\resultalt}[2]{\makebox[3em]{\hfill#1}\makebox[3.2em]{\hfill(#2)}}
\newcommand{\resultboldalt}[2]{\makebox[3em]{\hfill\textbf{#1}}\makebox[3.2em]{\hfill\textbf{(#2)}}}
\newcommand{\resultblueitalt}[2]{\makebox[3em]{\hfill\emph{\color{blue} #1}}\makebox[3.2em]{\hfill\emph{\color{blue} (#2)}}}

\newtheorem{theorem}{Theorem}[section]

\usepackage[capitalize,nameinlink]{cleveref}
\newcommand{\acro}[1]{\textsc{\MakeLowercase{#1}}}  % lower case acronyms
\DeclareMathOperator*{\argmax}{arg\,max}
\usepackage{dsfont}  % bb 1
\Crefname{section}{Sect.}{Sects.}
\Crefname{equation}{Eq.}{Eqs.}
\Crefname{figure}{Fig.}{Figs.}
\Crefname{tabular}{Tab.}{Tabs.}
\Crefname{table}{Tab.}{Tabs.}
\Crefname{appendix}{Appx.}{Appxs.}
\Crefname{algorithm}{Alg.}{Algs.}
\Crefname{observation}{Obs.}{Obs.}

\theoremstyle{observation}
\newtheorem{observation}[theorem]{Observation}

\everypar{\looseness=-1}

\usepackage{tabularx}  % equal-column-width tables
\usepackage{multirow}  % vertically merged cells
\usepackage{pgfplots}  % tikz
\usepgfplotslibrary{groupplots,dateplot}
\pgfplotsset{compat=1.17}

\pgfplotsset{
    legend image with text/.style={
        legend image code/.code={%
            \node[anchor=center] at (0.3cm,0cm) {#1};
        }
    },
}

\usepackage{array}

\newcolumntype{x}[1]{%
>{\centering\hspace{0pt}}p{#1}}%

\usepackage{algorithmic}
\usepackage{algorithm}

\usepackage{graphics}

\pdfoutput=1

\begin{document}

% If your paper is accepted and the title of your paper is very long,
% the style will print as headings an error message. Use the following
% command to supply a shorter title of your paper so that it can be
% used as headings.
%
\runningtitle{Nonmyopic Multiclass Active Search with Diminishing Returns for Diverse Discovery}

% If your paper is accepted and the number of authors is large, the
% style will print as headings an error message. Use the following
% command to supply a shorter version of the authors names so that
% they can be used as headings (for example, use only the surnames)
%
%\runningauthor{Surname 1, Surname 2, Surname 3, ...., Surname n}

\twocolumn[

\aistatstitle{Nonmyopic Multiclass Active Search with Diminishing Returns\\ for Diverse Discovery}

\aistatsauthor{Quan Nguyen \qquad Roman Garnett}

\aistatsaddress{ \\Washington University in St.\ Louis  } ]

\begin{abstract}
Active search is a setting in adaptive experimental design where we aim to uncover members of rare, valuable class(es) subject to a budget constraint.
An important consideration in this problem is diversity among the discovered targets -- in many applications, diverse discoveries offer more insight and may be preferable in downstream tasks.
However, most existing active search policies either assume that all targets belong to a common positive class or encourage diversity via simple heuristics.
We present a novel formulation of active search with multiple target classes, characterized by a utility function chosen from a flexible family whose members encourage diversity via a diminishing returns mechanism.
We then study this problem under the Bayesian lens and prove a hardness result for approximating the optimal policy for arbitrary positive, increasing, and concave utility functions.
Finally, we design an efficient, nonmyopic approximation to the optimal policy for this class of utilities and demonstrate its superior empirical performance in a variety of settings, including drug discovery.
\end{abstract}

\section{Introduction}

A theme underlying many real-world applications is the need to rapidly discover rare, valuable instances from massive databases in a budget-efficient manner.
For example, both drug discovery and fraud detection entail searching through huge spaces of candidates for rare instances exhibiting desired properties: binding activity against a biological target for drug discovery, or fraudulent behavior for fraud detection.
In both of these cases, labeling a given data point is expensive: synthesis and characterization in a laboratory for drug discovery, and human intervention (and possibly lost sales) for fraud detection.
This cost of labeling rules out exhaustive scanning and raises a challenge in experimental design.
The \emph{active search} (\acro{AS}) framework
frames such tasks in terms of active learning, where one iteratively queries an expensive oracle to determine
% that returns a label indicating the class membership of a chosen datapoint.
whether chosen data points are valuable.
%The need for label-efficiency arises in use cases where obtaining a label from the oracle is expensive, limiting the number of queries that may be made.
The goal of \acro{AS} is to adaptively design queries for labeling in order to identify as many valuable data points as possible under a given budget.

\acro{AS} has enjoyed a great deal of study \citep{garnett2012bayesian,jiang2017efficient,jiang2018efficient,jiang2019cost,nguyen2021nonmyopic},
and strong theoretical results and efficient algorithms are known.
Most relevant to this work, \citet{jiang2017efficient} established a strong hardness result for \acro{AS}.
Namely, \emph{no} computationally tractable policy (i.e., one that runs in polynomial time with respect to its querying budget) can guarantee recovery within any constant factor of the (exponential-time) optimal policy in the worst case.
This proof was by an explicit construction of arbitrarily hard problems.
Nonetheless, \citet{jiang2017efficient} were able to develop an efficient, nonmyopic policy that achieves impressive empirical results.

Prior work on \acro{AS} has operated in a binary setting where every data point is either valuable or not.
The total number of discoveries made in a given budget is then used as a utility function during experimental design, encoding equal marginal utility for every discovery made.
% However, this preference model does not apply to many practical scenarios where the utility is nonlinear.
% For example, having identified many members of a class could \emph{decrease} one's utility for obtaining yet another member of the same class, such as the case in scientific discovery where novel discoveries are more valuable.
% Motivated by this natural preference model of diminishing returns, we study the \acro{AS} problem with utility on the log scale of the number of queried targets.
% In the binary case (where there is one target class), this induces a \emph{risk-averse} utility model.
% In the more general multiclass case, this function assigns more value to discoveries of novel classes that have not been frequently observed, thus encouraging diversity within the collected data set.
However, this may not adequately capture preferences over experimental outcomes in many practical scenarios, where there may be \emph{diminishing returns} in finding additional members of a frequently observed class.
This is often the case in, e.g., scientific discovery, where a discovery in a novel region of the design space may offer more marginal insight than the 100th discovery in an already densely labeled region.
In this work, we will consider a \emph{multiclass} variant of the \acro{AS} problem, wherein discoveries in rare classes are awarded more marginal utility than those in an already well-covered classes.
As we will see, this approach naturally encourages \emph{diversity} among the points discovered.

After defining this problem, we study it through the lens of Bayesian decision theory.
%and make two main contributions.
% First, we extend the hardness of approximation proof by \citet{jiang2017efficient} to our setting, which we show to be applicable even in the binary case.
% Second, we propose an approximation to the optimal policy that is also efficient and nonmyopic.
% We demonstrate that similar to nonmyopic policies from previous work, ours can leverage budget-awareness to dynamically balance between exploration and exploitation.
% Further, our policy either takes on a risk-averse behavior in the binary case, or effectively biases for diversity in the multiclass case (an example of this is shown in \cref{fig:square}).
% A wide range of experiments show that our policy is empirically superior to many benchmarks in the literature by a large margin.
We begin by outlining how we can capture the notion of diminishing returns in marginal discovery through an (arbitrary) positive, increasing, and concave utility function.
%Subject to these constraints, we leave the details of the utility function to be determined by the needs of a particular setting
We then extend the hardness of approximation result by \citet{jiang2017efficient} from the linear utility to this much larger family, demonstrating that search is fundamentally difficult for a broad range of natural utility functions.
We then propose an approximation to the optimal policy for problems in this this class that is both computationally efficient and nonmyopic.
We show that the resulting algorithms effectively encourage diversity among discoveries, and, similar to nonmyopic policies from previous work, leverage budget-awareness to dynamically balance exploration and exploitation.
We demonstrate the superior empirical performance of our approach through an exhaustive series of experiments,
including in a challenging drug discovery setting.
Across the board, our proposed policy recovers both better balanced and richer data sets than a suite of strong baselines.
%In short, we: (1) define a novel and practically relevant search setting, (2) derive an efficient nonmyopic policy entailing novel approximation/analysis in its construction, (3) a theoretical hardness result regarding approximating the optimal policy, and (4) experimental validation using data from drug discovery, a target application area.

% The paper is organized as follows.
% We introduce the \acro{AS} problem with diminishing returns and derive the Bayesian optimal policy in \cref{sec:definition}.
% \cref{sec:approximation} details our proposed policy that nonmyopically approximates the optimal policy.
% \cref{sec:related} discusses related 

%\setcounter{section}{1}
\section{Problem Definition}
\label{sec:definition}

\looseness=-1
We first introduce the multiclass active search problem with diminishing returns and present the Bayesian optimal policy.
This policy will be hard to compute (or even approximate), but will inspire an approximation developed in the next section.
%, which is further shown to be hard to approximate by any constant factor in polynomial time.
% \subsection{Active Search with Diminishing Returns}
Suppose we are given a large but finite set of points $\mathcal{X} \triangleq \{ x_i \}$, each of which belongs to exactly one of $C$ classes, denoted by $[C] \triangleq \{ 1, 2, \ldots, C \}$, where $C > 2$.
We assume class-$1$ instances are abundant and uninteresting, while other classes are rare and valuable; we call the members of these classes \emph{targets}.
The class membership of a given point $x \in \mathcal{X}$ is not known \emph{a priori} but can be determined by making a query to an oracle that returning its label $y = c$.
We assume this labeling procedure is expensive and can only be accessed a limited number of times $T \ll n \triangleq \vert \mathcal{X} \vert$ -- the querying budget.
Denote a given data set of queried points and labels as $\mathcal{D} = \{ (x_i, y_i) \}$, and $\mathcal{D}_t$ as the data set collected after $t$ queries to the oracle in a given search.

\begin{figure}[t]
\begin{center}
\hfill
\includegraphics[width=0.5\linewidth]{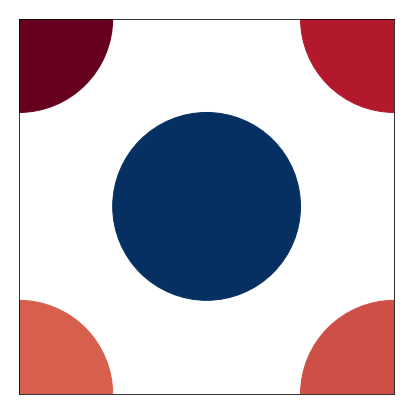}
\hfill
% This file was created with tikzplotlib v0.9.17.
\begin{tikzpicture}

\begin{axis}[
width=60,
height=151.5,
axis y line*=right,
tick align=outside,
xmin=0, xmax=0.1,
xmajorticks=false,
y grid style={white!69.0196078431373!black},
ymin=0, ymax=1,
ytick pos=right,
ytick style={color=black},
ytick={0,0.25,0.5,0.75,1},
yticklabel style={font=\scriptsize,anchor=west},
yticklabels={-50\%,-25\%,0\%,25\%,50\%}
]
\path [draw=white, fill=white, line width=0.004pt]
(axis cs:0,0)
--(axis cs:0,0.00390625)
--(axis cs:0,0.99609375)
--(axis cs:0,1)
--(axis cs:0.1,1)
--(axis cs:0.1,0.99609375)
--(axis cs:0.1,0.00390625)
--(axis cs:0.1,0)
--(axis cs:0.1,0)
--cycle;
\addplot graphics [includegraphics cmd=\pgfimage,xmin=0, xmax=0.1, ymin=0, ymax=1] {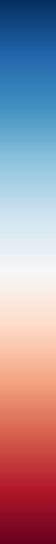};
\end{axis}

\end{tikzpicture}
\hfill
\caption{
Discoveries by region relative to a uniform target distribution by the state-of-the-art policy \acro{ENS}, in a toy search problem where points within the visible regions in the unit square are considered search targets.
\acro{ENS} over-samples the center region, the most common and easiest-to-identify target class, and collects a highly unbalanced data set.
}
\label{fig:square_ens}
\end{center}
\end{figure}

Our high-level goal is to design a policy that decides which elements of $\mathcal{X}$ should be queried in order to uncover as many targets as possible.
Preferences over different data sets (experimental outcomes) are expressed via a \emph{utility function;} previous work \citep{garnett2012bayesian,jiang2017efficient} has used a linear utility in the binary setting $C = 2$:
\begin{equation}
u \left( \mathcal{D} \right) = \sum\nolimits_{y \in \mathcal{D}} \mathds{1} \{ y > 1 \},
\end{equation}
\looseness=-1
which effectively groups all targets in a common positive class and assigns equal reward to each discovery.
In many practical scenarios, however, once a target class has been thoroughly investigated, the marginal utility of finding yet more examples decreases and we would prefer to either expand a rarely sampled class or discover a new one.
For example, in drug discovery -- one of the main motivating applications for \acro{AS} -- screening procedures optimized for hit rate tend to propose very structurally similar compounds and lead to an overall decline in usefulness of these discoveries downstream \citep{galloway2010diversity}.
This has lead to efforts to artificially encourage diversity when generating new screening experiments, as a way to induce the desired search behavior \citep{benhenda2017chemgan,pereira2021diversity}.

The preferences above reflect the notion of \emph{diminishing returns.}
We propose to capture diminishing returns for marginal discoveries in a known class $c$ (and thereby encourage diversity in discoveries) with a reward function $f_c$:
\begin{align}
u \left( \mathcal{D} \right) & \triangleq \sum_{c > 1} f_c \Big( \sum_{y \in \mathcal{D}} \mathds{1} \{ y = c \} \Big)
= \sum_{c > 1} f_c \left( m_c \right) \label{our_utility},
\end{align}
where $f_c$ is a positive, increasing, and concave function quantifying our reward given the number of found targets from a given class $c$.
The term $m_c$ denotes the number of targets of class $c$ in data set $\mathcal{D}$;
we also use $m_{c, t}$ to denote the corresponding number in $\mathcal{D}_t$ at time $t$.
% We note that any utility encoding decreasing marginal gains (that is, concave) could be an appropriate choice for this setting, and our proposal of the log is meant as a reasonable default when no better alternative is available.
% We will see later in \cref{sec:approximation} that the key element of our algorithm is valid with any other concave utility.
Any utility encoding decreasing marginal gains (that is, concave) is an appropriate choice for this setting, and we will see later in \cref{sec:approximation} that the key element of our algorithm is valid with any concave utility.
In our experiments, we use the logarithmic function $f_c(x) = \log (x + 1)$ (which has seen a wide range of applications,
namely modeling utility of wealth \citep{bernoulli1954exposition} or production output \citep{pemberton2015mathematics} in economics)
as a reasonable default with which to generate our main results.
We also present results showing that our methods generalize well to other possibilities such as the square root utility $f_c(x) = \sqrt{x}$.

Under this model, the marginal gain of an additional discovery decreases with the size of the corresponding class in $\mathcal{D}$.
% When the concave reward function $f_c$ is the same across all target classes, the utility function effectively rewards maximum label diversity within the collected targets: across different data sets $\mathcal{D}$ with the same number of targets, utility is maximized when the target counts across the valuable classes are equal, i.e., $m_2 = \cdots = m_C$.
Consider a toy problem illustrated in \cref{fig:square_ens}, where we wish to search for points close to the center and corners of the unit square, each representing a target class.
The state-of-the-art \acro{ENS} policy \citep{jiang2017efficient}, which uses a linear utility, recovers many targets but over-exploits the center region.
% thus scoring low on our diversity-informed utility measure.
This is undesirable behavior when we would prefer to have balanced discoveries across all classes; ideally, we would like to achieve a uniform target distribution (an equal number of hits across all classes for maximum diversity), where all five regions in the plot are transparent. 
We will develop policies that can achieves such balance in the next section.

\subsection{The Bayesian Optimal Policy}

\looseness=-1
We now derive the optimal policy in the expected case using Bayesian decision theory.
We first assume access to a model computing the posterior probability that a point $x \in \mathcal{X}$ belongs to class $c \in [C]$ given an observed data set $\mathcal{D}$, denoted by $p_c \left( x \mid \mathcal{D} \right) \triangleq \Pr \left( y = c \mid x, \mathcal{D} \right)$.
(We sometimes omit the dependence on $\mathcal{D}$ in the notation when the context is clear.)
%We describe our implementation of a $k$-nearest neighbor used in our experiments in \cref{subsec:model}, but note that the following discussion allows the use of any such model.
This model may be arbitrary.
Now, suppose we are currently at iteration $t + 1 \leq T$, having collected data set $\mathcal{D}_t$, and now need to identify the next point $x_{t + 1} \in \mathcal{X} \setminus \mathcal{D}_t$ to query the oracle with.
The optimal policy selects the point that maximizes the expected utility of the \emph{terminal} data set $\mathcal{D}_T$, conditioned on the current query, recursively assuming that future queries will also be made optimally:
\begin{equation}
\label{eq:optimal}
x_{t + 1}^* = \argmax_{x_{t + 1} \in \mathcal{X} \setminus \mathcal{D}_t} \mathbb{E} \bigl[ u \left( \mathcal{D}_T \right) \mid x_{t + 1}, \mathcal{D}_t \bigr].
\end{equation}
This expected optimal utility may be computed using backward induction \citep{bellman1957dynamic}.
In the base case where $t = T - 1$ and we are faced with the very last query $x_T$,
% \begin{multline*}
% \mathbb{E} \bigl[ u \left( \mathcal{D}_T \right) \mid x_T, \mathcal{D}_{T - 1} \bigr] \\
% = \sum_{c \in [C]} u \left( \mathcal{D}_T \right) p_c(x_T) % \Pr \left( y_T \mid x_T, \mathcal{D}_{T - 1} \right).
% \end{multline*}
\begin{equation}
\mathbb{E} \bigl[ u \left( \mathcal{D}_T \right) \mid x_T, \mathcal{D}_{T - 1} \bigr] = \sum_{c \in [C]} u \left( \mathcal{D}_T \right) p_c \left( x_T \mid \mathcal{D}_{T - 1} \right).
\end{equation}
Maximizing this expectation is equivalent to maximizing the expected \emph{marginal utility gain}
\begin{multline}
\label{eq:one_step}
\Delta \left( x_T \mid \mathcal{D}_{T - 1} \right) \triangleq \sum_{c > 1} p_c \left( x_T \mid \mathcal{D}_{T - 1} \right) \times \\
\Big[ f_c \left( m_{c, T - 1} + 1 \right) - f_c \left( m_{c, T - 1} \right) \Big].
\end{multline}
For each class $c$,
this quantity not only increases as a function of the positive probability $p_c$, but also decreases as a function of the number of targets already found in that class.
Therefore, even at this very last step, the optimal decision balances between hit probability and discovery/extension of a rare class.
When more than one query remains in our budget, the expected optimal utility in \cref{eq:optimal} expands into
\begin{multline}
\mathbb{E} \bigl[ u \left( \mathcal{D}_T \right) \mid x_{t + 1}, \mathcal{D}_t \bigr] = u \left( \mathcal{D}_t \right) + \Delta \left( x_{t + 1} \mid \mathcal{D}_t \right) + \\
\mathbb{E}_{y_{t + 1}} \Big[ \max_{x_{t + 2}} \mathbb{E} \bigl[ u \left( \mathcal{D}_T \right) \mid x_{t + 2}, \mathcal{D}_{t + 1} \bigr] - u \left( \mathcal{D}_{t + 1} \right) \Big],
\end{multline}
where $\mathbb{E} \bigl[ u \left( \mathcal{D}_T \right) \mid x_{t + 2}, \mathcal{D}_{t + 1} \bigr]$ is the expected utility that is a step further into the future and may be recursively computed using the same expansion.
Here, we note while the first term in the sum on the right-hand side (the utility at the current step $u \left( \mathcal{D}_t \right)$) is a constant, the other two terms may be interpreted as balancing between exploitation from the immediate reward (the marginal gain $\Delta \left( x_{t + 1} \mid \mathcal{D}_t \right)$), and exploration from the future rewards to be optimized by subsequent queries (the expected future utility).
Overall, computing this expectation involves $(\ell - 1)$ further nested expectations and maximizations, where $\ell = T - t$ is the search horizon.
This has a time complexity of $O \left( (C \, n)^\ell \right)$, making finding the optimal decision intractable for any large data set.

A potential solution to this problem is to limit the lookahead horizon by pretending that $\ell$ is small, thus myopically approximating the optimal policy.
The simplest form of this is to set $\ell = 1$ and greedily optimize for the one-step expected marginal utility gain in \cref{eq:one_step}.
We refer to the resulting policy as the \emph{one-step} policy.
Since our utility function has elements with diminishing returns, a question naturally arises as to whether the results from the submodularity \citep{krause2007near} and adaptive submodularity \citep{golovin2011adaptive} literature apply here, and whether the greedy strategy of the one-step policy could approximate the optimal policy well.
In the next subsection, we present the perhaps surprising result that \emph{no} polynomial-time policy can approximate the optimal policy by any constant factor.

\subsection{Hardness of Approximation}

\looseness=-1
Assuming access to a unit-cost conditional probability $p_c \left( x \mid \mathcal{D} \right)$ for any point $x \in \mathcal{X}$ and data set $\mathcal{D}$, we obtain the same hardness result of \citet{jiang2017efficient} for the broad range of utility functions considered here:%
\begin{theorem}\label{thm:hardness}
There is no polynomial-time policy providing any constant factor approximation to the optimal expected utility in the worst case.%
\end{theorem}%
%\vskip-1.5ex
Our proof strategy follows that of \citet{jiang2017efficient}.
We construct a family of hard problem instances, where in each instance a secret set of points encodes the location of a larger ``treasure'' of targets.
The probability of discovering this treasure is extremely small without observing the secret set first, which in itself is vanishingly unlikely to happen in polynomial time.
Further, the average hit rate outside of the treasure set is vanishingly low, making it infeasible to compete with the optimal policy.
Remarkably, we can construct such hard problem instances for any
utility that is
positive, increasing, unbounded, and concave in the number of discoveries in each class.
The complete proof is included in \cref{sec:hardness}.

Despite this negative result, we hope to design \emph{empirically} effective policies.
% Previous work \citep{garnett2012bayesian,jiang2017efficient,jiang2019cost} has demonstrated the benefits of nonmyopia, that is, reasoning about the utility of a given query in terms of both immediate reward and future progress, in \acro{AS} under linear utility both theoretically and empirically.
% We aim to do the same here, moving away from the greedy strategy and employing a nonmyopic policy.
Previous work has demonstrated that nonmyopic policies offer both theoretical and empirical benefits when working with the linear utility function, and that budget-awareness is in particular can be especially beneficial for a policy to effectively balance exploration and exploitation.
In the next section, we propose an efficient, nonmyopic approximation to the optimal policy for the class of utility functions we consider here, which we will show later also improves practical performance.

\section{Efficient Nonmyopic Approximation}
\label{sec:approximation}

We propose a batch rollout approximation to the optimal policy similar to the \acro{ENS} algorithm for binary \acro{AS} \citep{jiang2017efficient} and the \acro{GLASSES} algorithm for Bayesian optimization \citep{gonzalez2016glasses}.
The key idea is to assume that after a proposed query in the current iteration, all remaining budget will be spent \emph{simultaneously} on a single batch of queries exhausting the budget.
This assumption simplifies the decision tree we must analyze, reducing its depth to $2$ while expanding the branching factor of the last layer.
Under the linear utility model, the expected marginal utility of a final batch of queries conveniently decomposes into a sum of positive probabilities of individual batch members.
The optimal final batch therefore consists of the points with the highest probabilities, which may be computed efficiently.
By matching the size of the following batch to the number of queries remaining, we can effectively account for our remaining budget when computing the expected utility of a given putative query.
Unfortunately, the linear decomposition enabling rapid computation in \acro{ENS} does not hold in our setting due to our nonlinear utility \eqref{our_utility}, and designing an effective batch policy requires more care.

\subsection{Making a Batch of Queries}

We first temporarily consider the subproblem of designing a batch of $b$ queries $X$ given a data set $\mathcal{D}$ to maximize the expected utility of the combined observation set, i.e., $\mathbb{E}_Y \big[ u \left( \mathcal{D} \cup X, Y \right) \big]$,
where the expectation is taken over $Y$,%
\footnote{In this section, expectations over $Y$ are universally conditioned on knowledge of $X$ and $\mathcal{D}$; we drop this conditioning from the expressions to clarify the main ideas.}
the labels of  $X$.
As labels may be conditionally dependent and the utility function $u$ is not linear, exact computation of this expectation requires iterating over all $C^b$ realizations of the label set $Y$.
This is infeasible unless $b$ is very small, and represents the primary challenge we must overcome.

A crude but effective mechanism to address the conditional dependence of labels is to simply ignore the dependence (a ``mean field approximation'').
This relieves us from having to update the posterior for unseen points given fictitious observations arising in the computation.
%Under this assumption, \acro{ENS} becomes the \emph{optimal} policy given the linear utility.
However, in our setting, even if we assume conditional independence, we still face challenges in computing the expected utility:
\begin{equation}
\label{eq:exact}
%\begin{aligned}
\mathbb{E}_Y \big[ u \left( \mathcal{D} \cup X, Y \right) \big] 
%& = \mathbb{E}_Y \Big[ \sum_{c > 1} \log \left( 1 + m'_c \right) \mid X, \mathcal{D} \Big] \\
%&
= \sum_{c > 1}
\mathbb{E}_Y \Big[ f_c \left( m'_c \right) \Big],
%\mid X, \mathcal{D} \Big],
%\end{aligned}
\end{equation}
where $m'_c$ is the total number of targets belonging to class $c$ in the union set of $\mathcal{D}$ and a particular realization of $Y$.
In the interest of effective computation, we use Jensen's inequality to obtain an upper bound on the expected utility:
% \begin{align*}
% \mathbb{E}_Y \big[ u \left( \mathcal{D} \cup X, Y \right) \big] 
% & = \sum_{c > 1} \mathbb{E}_Y \Big[ f \left( m'_c \right) \Big] \\
% & \leq \sum_{c > 1} f \Big( \mathbb{E}_Y \left[ m'_c  \right]\Big).
% \end{align*}
\begin{equation}
\sum_{c > 1} \mathbb{E}_Y \Big[ f_c \left( m'_c \right) \Big] \leq \sum_{c > 1} f_c \Big( \mathbb{E}_Y \left[ m'_c  \right]\Big).
\end{equation}
Now, for a given class $c$, the inner expectation may be rewritten as the sum of probabilities (and some constants):
% \[
% \mathbb{E}_Y \left[ 1 + m'_c \mid X, Y, \mathcal{D} \right] = 1 + m_c + \sum_{x \in X} \Pr \left( y = c \mid x, \mathcal{D} \right).
% \]
\begin{equation}
\mathbb{E}_Y \left[ m'_c \right] = m_c + \sum_{x \in X} p_c \left( x \mid \mathcal{D} \right).
\end{equation}
% As such, we obtain an upper bound on the overall expected utility:
We then upper-bound the overall expected utility:
\begin{align}
\begin{split}
\mathbb{E}_Y \big[ u \left( \mathcal{D}\cup X, Y \right) \big] & \leq \overline{u}(X) \\
& \triangleq \sum_{c > 1} f_c \Big( m_c + \sum_{x \in X} p_c( x \mid \mathcal{D}) \Big).
\end{split}
\end{align}
We propose to use this upper bound, $\overline{u}$, to approximate the expected utility of a batch for the purposes of policy computation.
Here, we note that we may derive this bound for \emph{any} concave utility.
In \cref{sec:quality}, we present simulation results comparing the fidelity of this approximation to that of Monte Carlo sampling.
Overall, our method offers competitive accuracy even against sampling with a large number of samples of $Y$ (\textgreater 1000), while being significantly more computationally lightweight.
Here, speed is paramount since in batch rollout, computing the utility of a batch is a subroutine that needs to run many times ($C$ times for each putative query candidate, once for each putative label).
Another attractive feature of our approach is that $\overline{u}$ is a monotone submodular set function, which will facilitate efficient (approximate) maximization.

Our goal now is to find the batch $X$ that maximizes $\overline{u}$, as an approximation to the batch maximizing the expected one-step utility.
Na\"ively maximizing $\overline{u}$ requires iterating over $n \choose b$ candidate batches to compute the corresponding score.
However, we note that this score is a sum of concave, increasing functions, which are monotone submodular, and therefore $\overline{u}$ is a monotone submodular function itself.
We thus opt to \emph{greedily} optimize $\overline{u}$ by sequentially maximizing the pointwise marginal gain; the resulting batch provides a $(1 - 1 / e)$-approximation for the optimal batch \citep{nemhauser1978analysis,krause2007near}.
% This procedure is summarized in \cref{alg:batch}.
We briefly remark on the nature of the batches resulting from this greedy procedure, which naturally encourages batch members to be diverse in their likely labels: once a point having a high $p_c$ is added to $X$, others with high $p_{c'}$ for another target will be prioritized during the next selection.
This is a desideratum of a batch policy when seeking to encourage diversity, indicating the output of the algorithm is a reasonable approximation of the optimal batch.
Further, when $b = 1$ (that is, at the second-to-last iteration), this procedure makes the true expected-case optimal decision -- a reassuring feature.

\subsection{Completing the Algorithm}

\begin{algorithm}[tb]
   \caption{Diversity-aware active search (\acro{DAS})}
   \label{alg:DAS}
\begin{algorithmic}[1]
    \STATE {\bfseries inputs} observations $\mathcal{D}_t$, remaining budget $T - t$
    \STATE {\bfseries returns} $x^*_{t + 1}$ maximizing the score in \cref{eq:score}
    \FOR{$x_{t + 1} \in \mathcal{X} \setminus \mathcal{D}_t$}
        \FOR{$y_{t + 1} \leftarrow 1$ {\bfseries to} $C$}
            \STATE $\alpha \left( x_{t + 1} \mid y_{t + 1} \right) \leftarrow \overline{u} \left( X \mid \mathcal{D}_t \cup \{ (x_{t + 1}, y_{t + 1}) \} \right)$
        \ENDFOR
        \STATE $\alpha (x_{t + 1}) \leftarrow \sum_{c} p_c (x_{t + 1}) \, \alpha \left( x_{t + 1} \mid c \right)$
    \ENDFOR
    \STATE $x^*_{t + 1} \leftarrow \arg \max_{x_{t + 1}} \alpha (x_{t + 1})$
\end{algorithmic}
\end{algorithm}

With a method of constructing approximate one-step optimal batches in hand, we now complete our proposed policy, \emph{diversity-aware active search}, or \acro{DAS}.
For a candidate observation $x_{t + 1}$, we condition on each possible label $y_{t + 1} \in [C]$, approximate the optimal batch observation following $(x_{t + 1}, y_{t + 1})$, and average the resulting approximate terminal utility $\overline{u}$ over the labels $y_{t + 1}$:
\begin{equation}
\label{eq:score}
\alpha (x_{t + 1}) = \mathbb{E}_{y_{t + 1}} \big[ 
\overline{u} \left(X \right) \mid x_{t + 1}, \mathcal{D}_t \big],
\end{equation}
where $\overline{u}$ depends on the putative data $\mathcal{D} \cup (x_{t+1}, y_{t+1})$.
\acro{DAS} proceeds by selecting the candidate $x^*_{t + 1}$ that maximizes the score $\alpha$.
This procedure is summarized in \cref{alg:DAS}.

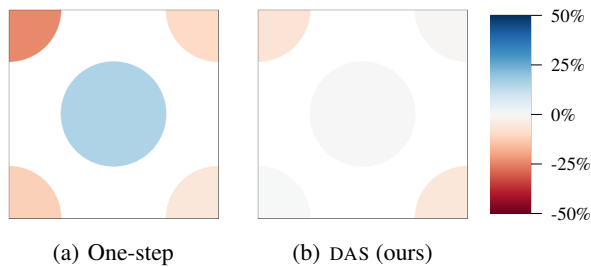
\begin{figure}[t]
\begin{center}
\hfill
\subfigure[One-step]{
\includegraphics[width=0.37\linewidth]{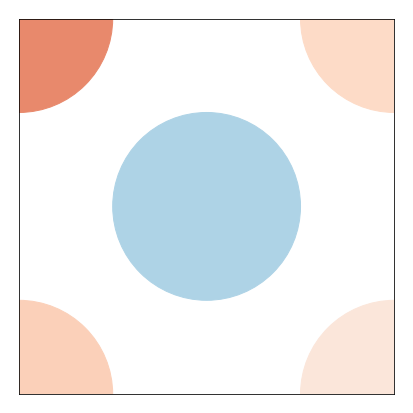}
}
\hfill
\subfigure[\acro{DAS} (ours)]{
\includegraphics[width=0.37\linewidth]{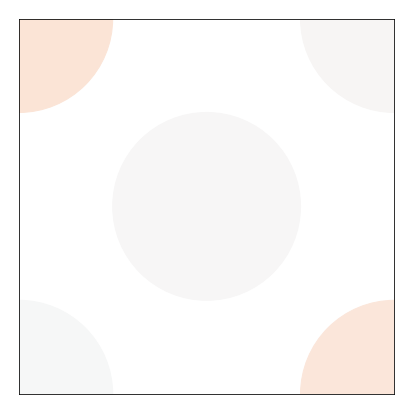}
}
\hfill
% This file was created with tikzplotlib v0.9.17.
\begin{tikzpicture}

\begin{axis}[
width=60,
height=120,
axis y line*=right,
tick align=outside,
xmin=0, xmax=0.1,
xmajorticks=false,
y grid style={white!69.0196078431373!black},
ymin=0, ymax=1,
ytick pos=right,
ytick style={color=black},
ytick={0,0.25,0.5,0.75,1},
yticklabel style={font=\scriptsize,anchor=west},
yticklabels={-50\%,-25\%,0\%,25\%,50\%}
]
\path [draw=white, fill=white, line width=0.004pt]
(axis cs:0,0)
--(axis cs:0,0.00390625)
--(axis cs:0,0.99609375)
--(axis cs:0,1)
--(axis cs:0.1,1)
--(axis cs:0.1,0.99609375)
--(axis cs:0.1,0.00390625)
--(axis cs:0.1,0)
--(axis cs:0.1,0)
--cycle;
\addplot graphics [includegraphics cmd=\pgfimage,xmin=0, xmax=0.1, ymin=0, ymax=1] {media/png/colorbar-000.png};
\end{axis}

\end{tikzpicture}
\hfill
\caption{
Discoveries by region relative to a uniform target distribution by the one-step policy and our proposal \acro{DAS}, in the problem visualized in \cref{fig:square_ens}.
One-step distributes queries more equally than the previously seen \acro{ENS}; however, center points are still over-represented.
\acro{DAS} constructs more diverse data sets and finds more rare corner targets.
}
\label{fig:square_rest}
\end{center}
\vskip -0.2in
\end{figure}

As mentioned, with the lookahead batch construction simulating future queries, \acro{DAS} accounts for not only the immediate reward but also the impact of the chosen query on future rewards.
Additionally, the latter quantity naturally decreases as a function of the remaining budget $b$, allowing our policy to be budget-aware and dynamically balance exploitation and exploration without any tradeoff parameter.
We briefly demonstrate the benefits of our approach by continuing with the example previously seen in \cref{fig:square_ens}, where \acro{ENS}, in seeking to maximize only recovery, collected highly unbalanced data sets.
\cref{fig:square_rest} shows the results of the one-step policy and \acro{DAS} under the same setting.
Compared to \acro{ENS}, one-step distributes its queries more equally, but it is our proposed policy \acro{DAS} that constructs the most diverse data set.

\subsection{Implementation}
\label{subsec:model}

A na\"ive implementation of the batch subroutine in \acro{DAS} has a complexity of $O( b \, n)$, and the entire \acro{DAS} procedure has a complexity of $O \big( C \, n \, (b \, n) \big) = O \left( C \, b \, n^2 \right)$, where again $n = \vert \mathcal{X} \vert$ is the size of our search space and $b$ is the remaining budget. %(i.e., the lookahead horizon).
% We discuss in this subsection strategies that allow our method to scale to large data sets.

We now describe the $k$-\acro{NN} model used in our experiments, introduced by \citet{garnett2012bayesian} in the binary setting.
The idea is to use the proportion of class-$c$ members among the observed nearest neighbors of a given point $x$ as the posterior marginal probability $p_c \left( x \mid \mathcal{D} \right)$.
Formally, denote the set of nearest neighbors of $x$ as $\acro{NN}(x)$ and the (potentially empty) subset of labeled neighbors as $\acro{LNN}(x) \subseteq \acro{NN}(x)$.
Then, the posterior probability of $x$ belonging to class $c$ is
\begin{equation}
p_c \left( x \mid \mathcal{D} \right) = \frac{\gamma_c + \sum_{x' \in \acro{LNN}(x)} \mathds{1} \{ y' = c \}}{1 + \vert \acro{LNN}(x) \vert},
\end{equation}
where each $\gamma_c \in (0, 1)$ is a hyperparameter acting as a ``pseudocount'', or our prior belief about the prevalence of class $c$ (since $p_c \left( x \mid \mathcal{D} \right) = \gamma_c$ if $\acro{LNN}(x) = \varnothing$). %; these pseudocounts add up to 1.
We detail our choice for this hyperparameter in \cref{sec:data}.

The $k$-\acro{NN} achieves reasonable generalization error in practice (in the sparsely labeled setting we are considering here), and can be rapidly updated given a new observation, which is a valuable feature with respect to our method.
Further, the $k$-\acro{NN} only uses the similarity matrix for $\mathcal{X}$, whose calculation only needs to be done once and can be accelerated by modern similarity search libraries such as Faiss \citep{johnson2019billion}.
This model also allows us to employ a branch-and-bound pruning strategy that speeds up the search for the query maximizing the score $\alpha$ at each iteration; we detail this procedure in \cref{sec:pruning}.

In the event where pruning is unsuccessful, we can sub-sample the space (e.g., subject to a user-specified cardinality constraint) and conduct the current search within the sampled pool.
This technique is extensively explored in \citet{mirzasoleiman2015lazier} and used by \citet{nguyen2021nonmyopic}.

\section{Related Work}
\label{sec:related}

Active search \citep{garnett2012bayesian} is a variant of active learning (\acro{AL}) \citep{settles2009active} where we aim not to learn an accurate model, but to collect members of rare and valuable classes.
Previous work has explored \acro{AS} under a wide range of settings, such as when the goal is to hit a targeted number of positives as quickly/cheaply as possible \citep{warmuth2002active,warmuth2003active,jiang2019cost}, when queries are made in batches \citep{jiang2018efficient}, or with multifidelity oracles \citep{nguyen2021nonmyopic}.
These studies all assumed there is only one target class, and collecting a target constitutes a constant reward.
Ours is the first to our knowledge to tackle multiclass \acro{AS}.
%using principled Bayesian decision theory.

Diversity as an objective has enjoyed great interest from the broader \acro{AL} community.
A common approach is to modify a typical \acro{AL} acquisition function to encourage diversity in the resulting queries.
For example, \citet{gu2015active} and \citet{yang2015multi} encouraged diversity by incorporating dissimilarity terms (computed via an \acro{RBF} kernel) into uncertainty sampling schemes.
\citet{brinker2003incorporating} used the angles between the hyperplanes induced by adding new points to the training set of an \acro{SVM}, and \citet{zhdanov2019diverse} considered the minimum distance between any pair of labeled points.
Others have employed coreset-based strategies \citep{sener2018active,agarwal2020contextual} to identify a set of diverse representative points.
Another popular strategy is to partition a given data set into different groups (e.g., using a clustering algorithm) and inspect the groups in a round-robin manner \citep{madani2004active,lin2018active,ma2020active,citovsky2021batch}.
We will apply a round-robin heuristic to a number of benchmarks in our experiments.

\citet{vanchinathan2015discovering} was motivated by a similar problem of uncovering a diverse, valuable subset.
Theirs is a regression setting in which diversity is measured in the feature space -- via the logdet of the Gram matrix of the collected data.
The proposed policy is myopic, maximizing the expected one-step marginal gain in a weighted sum of reward and diversity.
% and has two tradeoff parameters (for diversity and for exploration--exploitation) that need to be tuned.
\citet{he2007nearest} studied a related problem where the objective is to detect at least one instance of each rare target class as quickly as possible.
By assuming the target classes are highly concentrated, they design a policy that optimizes the difference in local density between a given point and its nearest neighbors, which is effective at identifying targets on the boundary.
% This strategy could be thought of as the one-step optimal policy when the utility is the number of target classes the collected data set spans.
\citet{malkomes2021beyond} considered the \emph{constraint active search} problem, in which they seek to find a diverse set of points satisfying a set of constraints.
The authors propose maximizing the expected improvement in a coverage measure given a new observation.
% The strategies employed by these studies are the one-step optimal policies under their respective utility functions.
% As we will see shortly, the analogous one-step policy in our setting is greatly outperformed by our nonmyopic proposal, demonstrating there is value in going beyond one-step lookahead.
% While the respective objectives of these policies are somewhat different from ours, we choose to include these policies as \acro{AL} baselines in our experiments and show that they are not competitive against our proposal.
We include these policies as \acro{AL} baselines in our experiments.

Closest to the motivation of our work, a line of research \citep{kothawade2021similar,kothawade2022prism} has explored a unified \acro{AL} framework for querying rare, diverse subsets of a large pool using submodular information measures.
Specifically, the neural network-based \acro{SIMILAR} algorithm \citep{kothawade2021similar} consists of \acro{AL} policies that can tackle problematic yet realistic learning scenarios such as imbalance in the training data, rare classes, out-of-distribution test data, and redundancy.
While it can be used for active search, \acro{SIMILAR} is not designed to specifically target discovery tasks; in \cref{sec:more_results}, we present results comparing the instantiation of \acro{SIMILAR} for rare class detection with our method, noting a slightly lower performance from \acro{SIMILAR}.
Further, our diversity-aware active search framework allows the utility function to be adjusted by the user to dynamically balance between discovery and diversity, a valuable feature in many discovery tasks.

Diversity has also been explored in the related task of Bayesian optimization \citep{garnett2022bayesian}.
A common approach is to incorporate a determinantal point process \citep{kulesza2012determinantal} to induce diversity in the feature space \citep{wang2018active,nava2021diversified}.
In the multiobjective setting, many policies leverage the diversity of their collected data in the Pareto space to design queries \citep{gupta2018exploiting,shu2020new,konakovic2020diversity}.

\section{Experiments}
\label{sec:experiments}

% \begin{table*}[t]
% \centering
% \begin{tabular}{ccccccc}
% \toprule
%  & \multirow{2}{*}{Quincunx} & \multicolumn{2}{c}{$\text{CiteSeer}^\text{x}$} & \multicolumn{3}{c}{Drug discovery} \\
% \cmidrule(lr){3-4} \cmidrule(lr){5-7}
%  &  & $C = 5$ & $C = 10$ & $C = 5$ & $C = 10$ & $C = 15$ \\
% \midrule
% \acro{AP-ENS} & 11.88 (0.40) & 16.57 (0.08) & 32.54 (0.50) & 10.29 (0.68) & 13.79 (0.85) &  \\
% \acro{RR}-greedy &  & 16.66 (0.14) & 33.08 (0.13) & 10.87 (0.82) & 18.66 (1.13) &  \\
% \acro{RR-UCB} &  & \begin{tabular}[c]{@{}c@{}}16.68 (0.12)\\ ($\beta^* = $)\end{tabular} & \begin{tabular}[c]{@{}c@{}}33.22 (0.13)\\ ($\beta^* = $)\end{tabular} & \begin{tabular}[c]{@{}c@{}}11.60 (0.90)\\ ($\beta^* = $)\end{tabular} & \begin{tabular}[c]{@{}c@{}}19.27 (1.13)\\ ($\beta^* = $)\end{tabular} & \begin{tabular}[c]{@{}c@{}}26.45 (0.96)\\ ($\beta^* = $)\end{tabular} \\
% \acro{RR-ENS} &  &  &  &  &  &  \\
% One-step & 13.14 (0.24) & 16.77 (0.13) & 34.01 (0.13) & 11.68 (0.92) & 19.06 (0.98) &  \\
% \acro{DAS} & 14.25 (0.07) & 17.37 (0.09) & 34.45 (0.11) & 13.34 (0.79) & 23.39 (0.83) & \\
% \bottomrule
% \end{tabular}
% \caption{}
% \label{tab:results}
% \end{table*}

We performed a series of experiments to evaluate the empirical performance of our policy \acro{DAS}.
% Experiments were performed on a small cluster built from commodity hardware comprising approximately 200 Intel Xeon \acro{CPU} cores, with approximately 10 \acro{GB} of \acro{RAM} available to each core.
% No \acro{GPU}s or other hardware was used in the computation.
As baselines, we considered related active learning/search algorithms \citep{he2007nearest,vanchinathan2015discovering,malkomes2021beyond} (see \cref{sec:related}), as well as the 
one-step lookahead policy, which greedily maximizes the marginal utility at each iteration.
Another baseline was \acro{ENS} \citep{jiang2017efficient}, the state-of-the-art for binary \acro{AS}, where we lumped all targets into a single positive class.

We also considered a family of policies that design queries in a \emph{round-robin} (\acro{RR}) manner.
In each iteration $t$, we choose a target class $c_t$ and seek to make a discovery for this target class.
A round-robin policy then continually rotates the target class among the positive classes throughout the search, devoting an equal amount of resources to each class.
The first of these policies is \acro{RR}-greedy, which for a given class index $c_t$ queries the point that maximizes the probability $p_{c_t}$.
Another round-robin baseline is \acro{RR-UCB}, which maximizes an upper confidence bound score \citep{auer2002using} corresponding to $c_t$: $p_{c_t} + \beta \sqrt{p_{c_t} (1 - p_{c_t})}$. Here $\beta$ is a hyperparameter trading off exploitation (class membership probability) and uncertainty (as measured by the standard deviation of the binary indicator $[y = c_t]$).
We evaluate this policy for $\beta \in \{ 0.1, 0.3, 1, 3, 10 \}$ and report the result of the best performing value of $\beta$, denoted $\beta^*$ in our results (\cref{tab:results}).
Finally, we consider \acro{RR-ENS}, which applies the \acro{ENS} heuristic to the subproblem of finding positives in the target class $c_t$.
The remaining budget is equally allocated among the positive classes, and we adjust the remaining budget when constructing lookahead batches accordingly.

\begin{table*}[t]
\centering
\caption{
Logarithmic utility and standard errors across 20 repeated experiments for each setting.
$C$ is the total number of unique classes in the search space.
The best performance in each column is highlighted in \textbf{bold}; policies not significantly worse than the best (according to a two-sided paired $t$-test with a significance level of $\alpha = 0.05$) are in \emph{\color{blue} blue italics}.
}
\vskip 10pt
\resizebox{\textwidth}{!}{\begin{tabular}{ccccccccc}
\toprule
 & \multicolumn{2}{c}{\acro{F}ashion-\acro{MNIST}} & Photoswitch & \multicolumn{2}{c}{$\text{CiteSeer}^\text{x}$} & \multicolumn{3}{c}{Drug discovery} \\
\cmidrule(lr){2-3} \cmidrule(lr){5-6} \cmidrule(lr){7-9}
 & easy & hard & & $C = 5$ & $C = 10$ & $C = 5$ & $C = 10$ & $C = 15$ \\
\midrule
\citet{he2007nearest} & \result{3.99}{0.16}	& \result{3.95}{0.16} & \result{12.65}{0.22} & \result{11.30}{0.11} & \result{24.61}{0.14} & \result{3.49}{0.23} & \result{7.19}{0.27} & \result{10.35}{0.35} \\
\citet{vanchinathan2015discovering} & \result{9.47}{0.50} & \result{6.99}{0.51} & \result{7.16}{0.25} & \result{16.28}{0.06} & \result{31.95}{0.49} & \result{10.19}{0.70} & \result{18.67}{0.88} & \result{25.51}{1.40} \\
\citet{malkomes2021beyond} & \result{11.31}{0.34} & \result{10.93}{0.47} & \result{6.39}{0.03} & \result{11.28}{0.01} & \result{22.16}{0.01} & \result{7.41}{0.31} & \result{13.77}{0.56} & \result{19.60}{0.66} \\
\midrule
\acro{ENS} & \result{10.48}{0.29} & \result{10.91}{0.14} & \result{5.04}{0.21} & \result{16.57}{0.08} & \result{32.54}{0.50} & \result{10.29}{0.68} & \result{13.79}{0.85} & \result{17.00}{1.15} \\
\midrule
\acro{RR}-greedy & \result{11.41}{0.36} & \result{10.41}{0.44} & \result{11.70}{0.30} & \result{16.66}{0.14} & \result{33.08}{0.13} & \result{10.87}{0.82} & \result{18.66}{1.13} & \result{24.87}{0.95} \\
\acro{RR-UCB} & \begin{tabular}[c]{@{}c@{}}\result{11.51}{0.37}\\ ($\beta^* = 1$)\end{tabular} & \begin{tabular}[c]{@{}c@{}}\result{11.28}{0.47}\\ ($\beta^* = 1$)\end{tabular} & \begin{tabular}[c]{@{}c@{}}\result{11.70}{0.30}\\ ($\beta^* = 3$)\end{tabular} & \begin{tabular}[c]{@{}c@{}}\result{16.68}{0.12}\\ ($\beta^* = 1$)\end{tabular} & \begin{tabular}[c]{@{}c@{}}\result{33.22}{0.13}\\ ($\beta^* = 3$)\end{tabular} & \begin{tabular}[c]{@{}c@{}}\result{11.60}{0.90}\\ ($\beta^* = 3$)\end{tabular} & \begin{tabular}[c]{@{}c@{}}\result{19.27}{1.13}\\ ($\beta^* = 3$)\end{tabular} & \begin{tabular}[c]{@{}c@{}}\result{26.45}{0.96}\\ ($\beta^* = 10$)\end{tabular} \\
\acro{RR-ENS} & \resultblueit{13.97}{0.09} & \resultbold{12.78}{0.31} & \result{12.54}{0.11} & \resultbold{17.47}{0.11} & \result{33.78}{0.16} & \result{10.56}{0.72} & \result{17.83}{0.66} & \result{23.58}{0.88} \\
\midrule
One-step & \result{11.83}{0.39} & \result{11.57}{0.47} & \result{8.67}{0.13} & \result{16.77}{0.13} & \result{34.01}{0.13} & \result{11.68}{0.92} & \result{19.06}{0.98} & \result{27.40}{1.11} \\
\acro{DAS} (ours) & \resultbold{14.02}{0.06} & \resultblueit{12.38}{0.30} & \resultbold{13.85}{0.27} & \resultblueit{17.37}{0.09} & \resultbold{34.45}{0.11} & \resultbold{13.34}{0.79} & \resultbold{23.39}{0.83} & \resultbold{31.48}{1.25} \\
\bottomrule
\end{tabular}}
\label{tab:results}
\end{table*}

\begin{figure}[t]
%\vskip 0.2in
%\begin{center}
\centering
\input{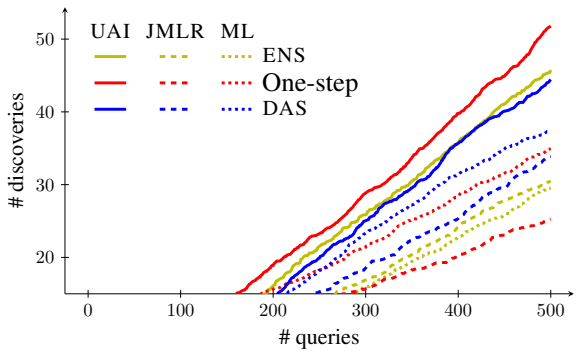}
\caption{
Average count of least-common papers found by different policies in the $\text{CiteSeer}^\text{x}$ $C = 10$ experiments.
\acro{DAS} finds more members of rare classes.
}
\label{fig:citeseer}
%\end{center}
\end{figure}

When relevant in the following experiments, we used a utility function that was logarithmic in marginal discoveries for each class:
% $f(x) = \log(x + 1)$.
$u \left( \mathcal{D} \right) = \sum_{c > 1} \log \left( 1 + m_c \right)$;
however, we also conducted a robustness study comparing performance for this utility function with an alternative that had significantly faster (square root) growth in each class.
We set our budget $T = 500$ unless specified otherwise and run each policy 20 times, each time with an initial training set containing  a randomly selected target.
% \subsection{The $\text{CiteSeer}^\text{x}$ data set}
% \label{subsec:citeseer}
We tested our policies on a wide range of data sets representing a diverse set of applications, which we briefly describe below.
More details on these data sets are included in \cref{sec:data}.

\textbf{Product recommendation.}
Our first task uses the \acro{F}ashion-\acro{MNIST} data set \citep{xiao2017fashion} of fashion item images to simulate a product recommendation setting.
Here, we assume a user is looking for specific classes of fashion articles while shopping online.
%That is, the user has a set of target classes, an instance of which we consider a ``hit''.
To simulate two different users (corresponding to two search problems), we select specific classes to be the products each user is looking for.
We sub-sample these classes uniformly at random, making the positives harder to uncover; the hit rate of a random policy is roughly 2\%.
While the $k$-\acro{NN} model has been found to perform well on the remaining data sets in previous studies \citep{garnett2012bayesian,jiang2017efficient,mukadum2021efficient}, with this data, we can select targets that are either easy or difficult to identify with the $k$-\acro{NN}, to simulate different levels of search difficulty and study the effect of the quality of the $k$-\acro{NN} on the performance of our methods.
To this end, we measure the predictive performance of the $k$-\acro{NN} and split the two search problems into an ``easy'' one, where the model scores highly on precision-at-$k$ metrics (particularly important in \acro{AS}), and a ``hard'' one, where the $k$-\acro{NN} scores lower.
More details are included in \cref{sec:data}.

\textbf{Photoswitch discovery.}
We also consider the task introduced by \citet{mukadum2021efficient} of searching for photoswitches (molecules that change their properties upon irradiation) in chemical databases that exhibit both desirable light absorbance and long half-lives.
% The data set contains 2049 molecules, 733 of which are targets.
Roughly 36\% of the molecules in the search space are targets.
In their study, the authors partitioned the points into 29 groups by their respective substructures, thus defining a multiclass \acro{AS} problem with $C = 30$ (a negative class and 29 positive classes).
% The number of targets in a class ranges from 0 to 121.
As this data set is smaller in size, we set the budget $T = 100$. %in each repeated experiment.

\textbf{The $\text{CiteSeer}^\text{x}$ data set.}
We use the $\text{CiteSeer}^\text{x}$ citation network data \citep{garnett2012bayesian}, which contains papers published at popular computer science conferences and journals.
The label of each paper is its publication venue, and our targets are machine learning and artificial intelligence proceedings.
We conduct two sets of experiments with different numbers of classes $C$.
For $C = 5$, we select papers from \acro{N}eur\acro{IPS}, \acro{ICML}, \acro{UAI}, and \acro{JMLR} as our four target classes (roughly a 14\% hit rate). %; an average of 3.5\% per class).
For $C = 10$, we further include \acro{IJCAI}, \acro{AAAI}, \acro{JAIR}, \emph{Artificial Intelligence,} and \emph{Machine Learning} (\acro{ML}) as targets, yielding a 31\% hit rate. % and a 3.5\% per class.

\begin{table*}[t]
\centering
\caption{
Average search utility and standard errors under various utility functions.
$C$ is the total number of unique classes in the search space.
The best performance in each column is highlighted in \textbf{bold}; policies that are not significantly worse than the best (according to a two-sided paired $t$-test with a significance level of $\alpha = 0.05$) are in \emph{\color{blue} blue italics}.
}
\vskip 10pt
\resizebox{0.85\textwidth}{!}{\begin{tabular}{ccccccc}
\toprule
 &  & utility function & One-step & \acro{DAS}$_\text{linear}$ (\acro{ENS}) & \acro{DAS}$_\text{sqrt}$ & \acro{DAS}$_\text{log}$ \\
\midrule
\multirow{3}{*}{CiteSeer$^\text{x}$} & \multirow{3}{*}{$C = 10$} & linear & \resultalt{445.90}{5.74} & \resultboldalt{459.25}{3.23} & \resultalt{448.35}{2.48} & \resultalt{433.55}{4.42} \\
 &  & sqrt & \resultalt{60.59}{0.46} & \resultalt{59.31}{0.81} & \resultboldalt{62.41}{0.24} & \resultalt{61.41}{0.33} \\
 &  & log & \resultalt{34.01}{0.13} & \resultalt{32.54}{0.50} & \resultboldalt{34.72}{0.09} & \resultalt{34.45}{0.11} \\
\midrule
\multirow{6}{*}{Drug discovery} & \multirow{3}{*}{$C = 10$} & linear & \resultalt{370.00}{18.37} & \resultboldalt{415.25}{12.05} & \resultalt{304.35}{22.73} & \resultalt{269.25}{18.47} \\
 &  & sqrt & \resultalt{36.33}{1.48} & \resultalt{32.12}{1.17} & \resultblueitalt{38.89}{1.76} & \resultboldalt{40.44}{1.66} \\
 &  & log & \resultalt{19.06}{0.98} & \resultalt{13.79}{0.85} & \resultblueitalt{21.07}{1.11} & \resultboldalt{23.39}{0.83} \\
\cmidrule(lr){2-7}
 & \multirow{3}{*}{$C = 15$} & linear & \resultalt{384.65}{11.08} & \resultboldalt{427.95}{12.88} & \resultalt{327.50}{14.82} & \resultalt{269.25}{18.47} \\
 &  & sqrt & \resultalt{46.60}{1.70} & \resultalt{36.71}{1.71} & \resultboldalt{55.50}{1.41} & \resultalt{49.20}{1.98} \\
 &  & log & \resultalt{27.40}{1.11} & \resultalt{17.00}{1.15} & \resultboldalt{34.03}{1.06} & \resultblueitalt{31.48}{1.25} \\
\bottomrule
\end{tabular}}
\label{tab:utility_misspec}
\end{table*}

\textbf{Drug discovery.}
Finally, we experiment with a drug discovery task using a massive chemoinformatic data set.
The goal is to identify chemical compounds that exhibit selective binding activity to a given protein.
The data set consists of 120 such activity classes from \acro{B}inding\acro{DB} \citep{liu2007bindingdb}.
For each class, there are a small number of compounds with significant binding activity -- these are our search targets.
In each  experiment for a given $C \in \{ 5, 10, 15 \}$, we select $(C - 1)$ of the 120 classes uniformly at random without replacement to form the targets.
They are then combined with 100\,000 ``drug-like'' entries in the \acro{ZINC} database \citep{sterling2015zinc}, which serve as the negatives, to make up our search space.
We note that each of these active classes has a unique structure and behavior, and combining multiple classes in one \acro{AS} problem makes ours a challenging task.
Here, the average prevalence of a target is 0.2\%; the hit rates are 1\%, 2\%, and 3\% for $C = 5$, $10$, and $15$, respectively.

\textbf{Discussion.}
We report the performance of the policies, quantified by the logarithmic utility function, in \cref{tab:results}.
Overall, \acro{DAS} either is the winner or does not perform significantly worse than the best policy.
This consistent performance highlights the benefits of our nonmyopic, budget-aware approach.
% One-step is usually the second-best among the baselines, demonstrating even a greedy strategy within our decision-theoretic framework is typically more effective than the other heuristics considered.
Under \acro{F}ashion-\acro{MNIST}, most methods work better on the easy problem than on the hard problem, showing the importance of the predictive model in \acro{AS}.
That said, our method remains competitive even under the harder setting.
Inspecting the optimal values for \acro{RR-UCB}'s tradeoff parameter $\beta^*$ in the $\text{CiteSeer}^\text{x}$ and drug discovery experiments, we notice a natural trend: as $C$ increases, so does the need for exploration, and larger values of $\beta$ are thus selected.

To illustrate \acro{DAS} is effective at constructing \emph{diverse} observations, we show in \cref{fig:citeseer} (whose $y$-axis is truncated for clarity) the average numbers of discoveries by the best three policies under the $\text{CiteSeer}^\text{x}$ $C = 10$ experiments for the three rarest classes: 
% \acro{UAI} (792 in total), \acro{JMLR} (739), and \acro{ML} (679) papers.
\acro{UAI}, \acro{JMLR}, and \acro{ML}.
Here, \acro{DAS} successfully finds more targets from the rarer classes of \acro{JMLR} and \acro{ML}, with a better balance among these classes as well.

\begin{figure}[t]
%\vskip 0.2in
\centering
\input{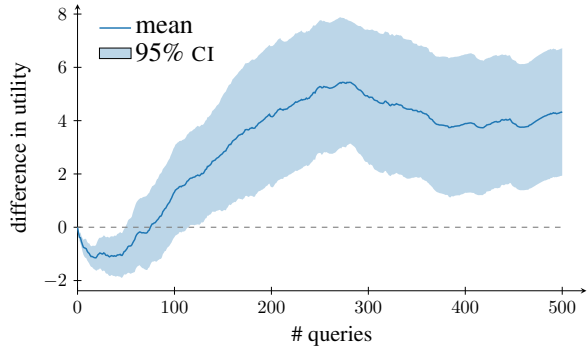}
\caption{
Difference in cumulative reward between \acro{DAS} and one-step across the drug discovery $C = 10$ experiments.
\acro{DAS} dynamically balances exploration and exploitation.
}
\label{fig:reward}
\end{figure}

By design, \acro{DAS} is always aware of its remaining budget during search and therefore dynamically balances exploration and exploitation.
We demonstrate this with the difference in the cumulative reward of \acro{DAS} vs.\ one-step under the drug discovery $C = 10$ setting in \cref{fig:reward}.
\acro{DAS} collects fewer rewards in the beginning while exploring the space.
As the search progresses, the policy transitions to more exploitation and ultimately outperforms the myopic one-step.
% This trend is consistent with what has been observed from \acro{ENS}-style policies in previous work \citep{jiang2017efficient,jiang2018efficient,nguyen2021nonmyopic}.

\textbf{Other utility functions and misspecification.}
One may reasonably ask whether \acro{DAS} still works under other possible utility functions, and how robust the method is against utility misspecification.
To tackle these questions, we reran the CiteSeer$^\text{x}$ $C = 10$ and drug discovery $C \in \{ 10, 15 \}$ experiments with the square root utility $u \left( \mathcal{D} \right) = \sum_{c > 1} \sqrt{m_c}$, which rewards additional discoveries of a known class more than the logarithm.
This presents an alternative utility with a different asymptotic behavior, but our method can be applied without any algorithmic modification.
\acro{DAS} again consistently performs the best across these settings (results in \cref{sec:more_results}), showing that the policy generalizes well to this utility.
% As another note on the flexibility of our framework, in addition to selecting the shape of the reward function, a user may even weight the target classes differently to prioritize certain classes, and \acro{DAS} can still run as-is.
As another note on the flexibility of our framework, a user may select different reward functions $f_c$ (e.g., by weighting the functions differently) to prioritize certain classes, and \acro{DAS} can still run as-is.

As for robustness against utility misspecification, we look for any performance drop when \acro{DAS} is evaluated under a utility different from what the policy uses during search.
First, we classify the variants of \acro{DAS} by the utility they use: (1) the version optimizing the logarithmic utility shown in \cref{tab:results} is denoted by \acro{DAS}$_\text{log}$, (2) the version using the square root utility is \acro{DAS}$_\text{sqrt}$, and (3) the version with the linear utility, \acro{DAS}$_\text{linear}$, reduces to \acro{ENS}.
We then evaluate these policies using all three utility functions.
We also include the one-step policy optimizing the correct utility in the results in \cref{tab:utility_misspec}.
Overall, each \acro{DAS} variant is competitive under the correct utility, always outperforming the corresponding one-step counterpart.
Crucially, both concave variants, \acro{DAS}$_\text{log}$ and \acro{DAS}$_\text{sqrt}$, perform well ``cross-utility'' in each other's setting, even outperforming \emph{the one-step policy with the correctly specified objective.}
Thus there is merit in adopting \acro{DAS} even when there may be uncertainty regarding the nuances of the user's ``true'' utility function.
We hypothesize this is because the concave utilities are similar in behavior -- both encourage diversity in the collected labels and are optimized by balanced data sets -- and a policy effective under one utility is likely to perform well under the other.
%This shows \acro{DAS}'s robustness against a misspecified concave utility.
The linear utility, on the other hand, does not exhibit diminishing returns and behaves differently from the others.
Utility misspecification here is thus more costly: \acro{DAS}$_\text{linear}$ is not competitive under concave utilities, and neither are the concave variants under linear utility.

\section{Conclusion}
\label{sec:conclusion}

We propose a novel active search framework that rewards diverse discoveries and study the problem from the Bayesian perspective.
We first prove a hardness result, showing the optimal policy cannot be approximated by a constant in polynomial time.
We then design a policy that simulates approximate optimal future queries in an efficient manner.
This nonmyopic planning allows our method to be aware of its remaining budget at any point during search and trade off exploitation and exploration dynamically.
Our experiments illustrate the empirical success of the proposed policy on real-world problems and its ability to build diverse data sets.

While many real-world applications are modeled by our multiclass \acro{AS} framework, our model assumes we know \emph{a priori} how many classes are present, which may be violated in many use cases.
In other applications, one might also consider the \emph{multilabel} setting where a data point can belong to more than one target class.
Investigating the problem under these settings is an interesting future work.
Another direction is to extend our approach to the batch setting where multiple queries run simultaneously.
% We note that while \acro{AS} is typically applied to scientific discovery problems, and this work itself presents no immediate negative societal impacts, the \acro{AS} framework is general and could potentially be used for nefarious purposes.

\subsubsection*{Acknowledgements}
We thank the anonymous reviewers for their feedback during the review stage.
This work was supported by the National Science Foundation (\acro{NSF}) under award numbers \acro{OAC}–1940224, \acro{IIS}–1845434, and \acro{OAC}–2118201.

\bibliography{main}
\bibliographystyle{plainnat}

\newpage
\appendix

\onecolumn
\section{Hardness of Approximation}
\label{sec:hardness}

\begin{figure*}[t]
\vskip 0.2in
\begin{center}
\hfill
\subfigure[Secret set $S$ of $2^m$ points]{
\raisebox{10pt}{\resizebox{0.4\linewidth}{!}{\input{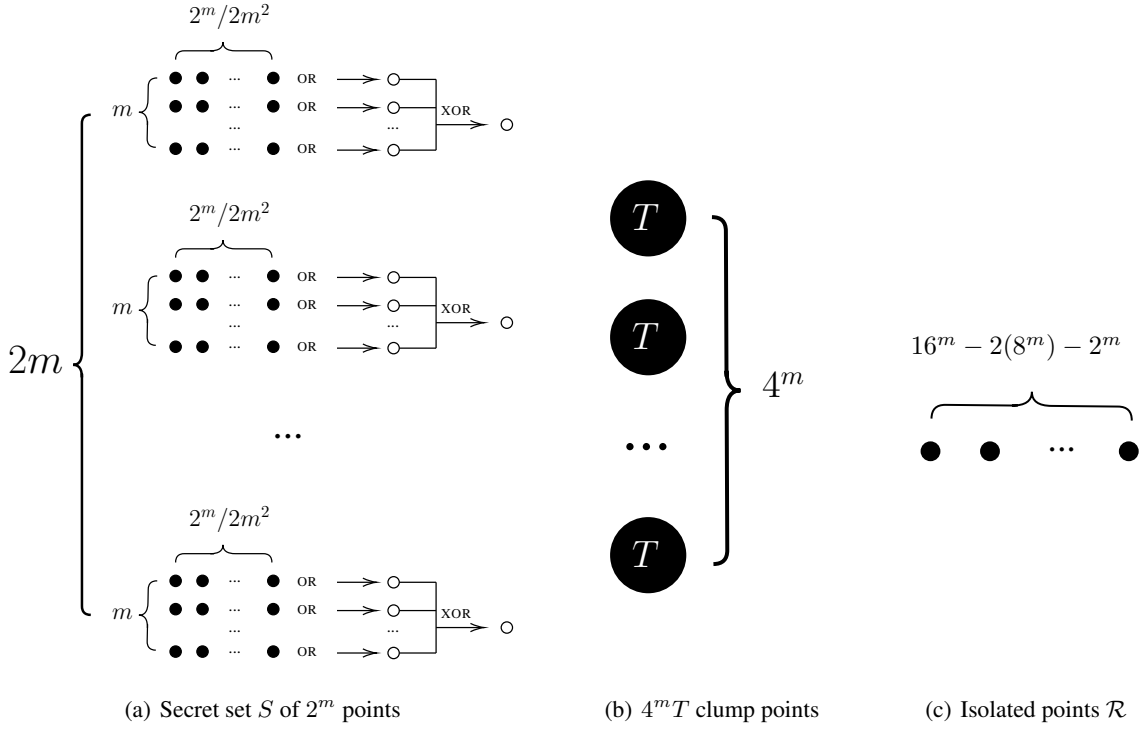}}}
\label{subfig:secret}
}
\hfill
\subfigure[$4^m T$ clump points]{
\raisebox{35pt}{\tikzset{every picture/.style={line width=0.75pt}} %set default line width to 0.75pt        

\begin{tikzpicture}[x=0.75pt,y=0.75pt,yscale=-1,xscale=1]
%uncomment if require: \path (0,1632); %set diagram left start at 0, and has height of 1632

%Shape: Brace [id:dp3764043694040142] 
\draw  [line width=1.5]  (216.54,958.81) .. controls (221.21,958.79) and (223.53,956.45) .. (223.51,951.78) -- (223.21,881.23) .. controls (223.18,874.56) and (225.49,871.22) .. (230.16,871.2) .. controls (225.49,871.22) and (223.15,867.9) .. (223.12,861.23)(223.13,864.23) -- (222.81,790.68) .. controls (222.79,786.01) and (220.45,783.69) .. (215.78,783.71) ;

% Text Node
\draw  [fill={rgb, 255:red, 0; green, 0; blue, 0 }  ,fill opacity=1 ]  (183.63, 784.33) circle [x radius= 18.72, y radius= 18.72]   ;
\draw (174.13,775.83) node [anchor=north west][inner sep=0.75pt]  [font=\Large]  {$\textcolor[rgb]{1,1,1}{T}$};
% Text Node
\draw (168.77,896.17) node [anchor=north west][inner sep=0.75pt]  [font=\Huge] [align=left] {...};
% Text Node
\draw (239.63,860.52) node [anchor=north west][inner sep=0.75pt]  [font=\Large]  {$4^{m}$};
% Text Node
\draw  [fill={rgb, 255:red, 0; green, 0; blue, 0 }  ,fill opacity=1 ]  (183.63, 844.33) circle [x radius= 18.72, y radius= 18.72]   ;
\draw (174.13,835.83) node [anchor=north west][inner sep=0.75pt]  [font=\Large]  {$\textcolor[rgb]{1,1,1}{T}$};
% Text Node
\draw  [fill={rgb, 255:red, 0; green, 0; blue, 0 }  ,fill opacity=1 ]  (183.63, 954.33) circle [x radius= 18.72, y radius= 18.72]   ;
\draw (174.13,945.83) node [anchor=north west][inner sep=0.75pt]  [font=\Large]  {$\textcolor[rgb]{1,1,1}{T}$};

\end{tikzpicture}}
\label{subfig:clumps}
}
\hfill
\subfigure[Isolated points $\mathcal{R}$]{
\raisebox{85pt}{\tikzset{every picture/.style={line width=0.75pt}} %set default line width to 0.75pt        

\begin{tikzpicture}[x=0.75pt,y=0.75pt,yscale=-1,xscale=1]
%uncomment if require: \path (0,1632); %set diagram left start at 0, and has height of 1632

%Shape: Circle [id:dp996525927117754] 
\draw  [fill={rgb, 255:red, 0; green, 0; blue, 0 }  ,fill opacity=1 ] (233,1213.5) .. controls (233,1211.01) and (235.01,1209) .. (237.5,1209) .. controls (239.99,1209) and (242,1211.01) .. (242,1213.5) .. controls (242,1215.99) and (239.99,1218) .. (237.5,1218) .. controls (235.01,1218) and (233,1215.99) .. (233,1213.5) -- cycle ;
%Shape: Brace [id:dp32804126490907715] 
\draw   (339,1198) .. controls (339.05,1193.33) and (336.74,1190.98) .. (332.07,1190.93) -- (298.57,1190.6) .. controls (291.9,1190.53) and (288.59,1188.17) .. (288.64,1183.5) .. controls (288.59,1188.17) and (285.24,1190.47) .. (278.57,1190.4)(281.57,1190.43) -- (245.07,1190.07) .. controls (240.4,1190.02) and (238.05,1192.33) .. (238,1197) ;
%Shape: Circle [id:dp06871766485296682] 
\draw  [fill={rgb, 255:red, 0; green, 0; blue, 0 }  ,fill opacity=1 ] (263,1213.5) .. controls (263,1211.01) and (265.01,1209) .. (267.5,1209) .. controls (269.99,1209) and (272,1211.01) .. (272,1213.5) .. controls (272,1215.99) and (269.99,1218) .. (267.5,1218) .. controls (265.01,1218) and (263,1215.99) .. (263,1213.5) -- cycle ;
%Shape: Circle [id:dp6616632709215684] 
\draw  [fill={rgb, 255:red, 0; green, 0; blue, 0 }  ,fill opacity=1 ] (333,1213.5) .. controls (333,1211.01) and (335.01,1209) .. (337.5,1209) .. controls (339.99,1209) and (342,1211.01) .. (342,1213.5) .. controls (342,1215.99) and (339.99,1218) .. (337.5,1218) .. controls (335.01,1218) and (333,1215.99) .. (333,1213.5) -- cycle ;

% Text Node
\draw (295,1210) node [anchor=north west][inner sep=0.75pt]  [font=\Large] [align=left] {...};
% Text Node
\draw (226,1153) node [anchor=north west][inner sep=0.75pt]    {$16^m - 2(8^m) - 2^m$};

\end{tikzpicture}}
\label{subfig:others}
}
\hfill
\caption{
An instance of \acro{AS} where any efficient algorithm can be arbitrarily worse than the optimal policy.
}
\label{fig:hardness_diagram}
\end{center}
\vskip -0.1in
\end{figure*}

We present the proof of \cref{thm:hardness}.
This is done by constructing a class of \acro{AS} problems similar to those described in \citet{jiang2017efficient} but with different parameterizations.
% We sketch out the main arguments here, and refer to the supplementary materials of \citet{jiang2017efficient} for a complete description.
Consider an \acro{AS} problem whose setup is summarized in \cref{fig:hardness_diagram} and described below.
The problem has $n = 16^m$ points, $C = 2^{m + 1} + 1$ classes ($2^{m + 1}$ positive classes), and the search budget is $T = 2^{m + 1}$, where $m$ is a free parameter.
We also have the reward function for each class $f_i$ to be the same (arbitrary) positive, increasing, unbounded, and concave $f$.
Each of the $n$ points is classified into two groups: ``clumps'' and ``isolated points.''

The former consists of  $4^m$ clumps, each of size $T$, and is visualized in \cref{subfig:clumps}.
All points within the same clump share the same label, and exactly one clump contains positives, each of which belongs to a different positive class.
In each instance of the problem class being described, this positive clump is chosen uniformly at random among the $4^m$ clumps.
As such, the prior marginal positive probability of any of these points is $p_{\text{clump}} = 4^{-m}$.

As for the isolated points, their labels are independent from one another.
The positives among the isolated points only belong to a single positive class.
The marginal probability of an isolated is set to be $p_{\text{isolated}} = 1 - 0.5^{\frac{2m^2}{2^m}}$.
These isolated points are further separated into two categories:
\begin{itemize}
\item A secret set $S$ of size $T / 2 = 2^m$, visualized in \cref{subfig:secret}, which encodes the location of the positive clump.
The set $S$ is first partitioned into $2m$ subsets $S_1, S_2, \ldots, S_{2m}$, each of size $2^m / 2m$.
Each subset $S_i$ encodes one virtual bit $b_i$ of information about the location of $S$, and is further split into $m$ groups of $2^m / 2m^2$ points, with each group encoding a virtual bit $b_{ij}$ by a logical \acro{OR}.
The aforementioned virtual bit $b_i$, on the other hand, is obtained via a logical \acro{XOR}: $b_i = b_{i1} \oplus b_{i1} \oplus \ldots \oplus b_{im}$.
\item The remaining points, denoted as $\mathcal{R}$ and visualized in \cref{subfig:others}, are completely independent from each other and any other points.
We have $\vert \mathcal{R} \vert = 16^m - 2(8^m) - 2^m$.
\end{itemize}

We first make the same two observations as in \citet{jiang2017efficient}.

\begin{observation}
\label{obs:obs1}
At least $m$ points from $S_i$ need to be observed in order to infer one bit $b_i$ of information about the location of the positive clump.
\end{observation}

Each $b_{ij}$ has the same marginal probability of being 1:
\[
\Pr \left( b_{ij} = 1 \right)
= 1 - \Pr \left( b_{ij} = 0 \right)
= 1 - (1 - p_{\text{isolated}})^{\frac{2^m}{2m^2}}
= 0.5.
\]

We also have $\Pr \left( b_i = 1 \right) = 0.5$, as the positive clump is chosen uniformly at random.
It is necessary to observe all virtual bits $b_{ij}$ from the same group $S_i$ to infer the bit $b_i$, since observing a fraction of the inputs of a \acro{XOR} operator does not change the marginal belief about the output $b_i$.
So, observing $(m - 1)$ or fewer points conveys no information about the positive clump.

\begin{observation}
\label{obs:obs2}
Observing any number of clump points does not change the marginal probability of any point in the secret set $S$.
\end{observation}

The knowledge of $b_i$ does not change the marginal probability of any $b_{ij}$.
This is to say no point in $S$ will have a different probability after observing $b_i$.
This means observing points outside of $S$ does not help distinguish $S$ from the remaining isolated points in $\mathcal{R}$.

With this setup, we now compare the performance of the optimal policy and the expected performance of a given polynomial-time policy.
To this end, we first consider the optimal policy with unlimited compute.
Before querying any point, the policy computes the marginal probability of an arbitrary fixed clump point, conditioning on observing every possible subset of the isolated points of size $m$ and fantasized positive labels.
This set of $O \left( n^m \right)$ inference calls will reveal the location of the secret set $S$, as only points in $S$ will update the probabilities of the fixed clump point.

Now the policy spends the first half of its budget querying $S$, the labels of which identify the positive clump.
The policy now spends the second half of the budget collecting these positive points.
The resulting reward, denoted as \acro{OPT}, is lower-bounded in the worst-case scenario where $S$ does not contain any positive point:
\[
\acro{OPT} \geq \frac{T}{2} \, f(1) = 2^m \, f(1).
\]

We now consider a policy $\mathcal{A}$.
Let $\alpha$ denote the total number of inference calls performed by $\mathcal{A}$ throughout its run.
At the $i^{\text{th}}$ inference call, $\mathcal{A}$ uses a training set $\mathcal{D}_i$ of size at most $T = 2^{m + 1}$.
We will show that $\mathcal{A}$ has a very small chance of collecting a large reward by considering several cases.

We first examine the probability that $\mathcal{A}$ finds the secret set $S$.
By \cref{obs:obs1,obs:obs2}, $\mathcal{A}$ cannot differentiate between the points in $S$ and those in $\mathcal{R}$ unless $\vert \mathcal{D}_i \cap S \vert \geq m$.
Suppose that before this inference call, the algorithm has no information about $S$ (which is always true when $i = 1$).
The chances of $\mathcal{A}$ choosing $\mathcal{D}_i$ such that $\vert \mathcal{D}_i \cap S \vert \geq m$ are no better than a random selection from $n - 2 \left( 8^m \right)$ isolated points.
We can upper-bound the probability of this event by counting how many subsets of size $2^{m + 1}$ would contain at least $m$ points from $S$ among all subsets of the $n - 2 \left( 8^m \right)$ isolated points:
\[
\Pr \left( \vert \mathcal{D}_i \cap S \vert \geq m \right) \leq \frac{{2^m \choose m} {n - 2(8^m) - m \choose 2^{m + 1} - m}}{{n - 2(8^m) \choose 2^{m + 1}}}.
\]

The \acro{RHS} may further be upper-bounded by considering
\[
\frac{{2^m \choose m} {n - 2(8^m) - m \choose 2^{m + 1} - m}}{{n - 2(8^m) \choose 2^{m + 1}}} = \frac{(2^m)! ~ (n - 2(8^m) - m)! ~ (2^{m + 1})!}{m! ~ (2^m - m)! ~ (2^{m + 1} - m)! ~ (n - 2(8^m))!},
\]
where
\begin{align*}
\frac{(2^m)!}{(2^m - m)!} & < (2^m)^m, \\\\
\frac{(2^{m + 1})!}{(2^{m + 1} - m)!} & < (2^{m + 1})^m, \\\\
\frac{(n - 2(8^m) - m)!}{m! ~ (n - 2(8^m))!} & < \frac{1}{(n - 2(8^m))^m}.
\end{align*}

The last inequality is due to
\[
\frac{(n - 2(8^m))^m}{m!} < \frac{(n - 2(8^m))!}{(n - 2(8^m) - m)!},
\]
which is true by observing that for each of the $m$ factors on each side,
\[
\frac{n - 2(8^m)}{i} < n - 2(8^m) - i + 1, \forall i = 1, \ldots, m.
\]

Overall, we upper-bound the probability that $\mathcal{A}$ hits $m$ points in $S$ with
\[
\Pr ( | \mathcal{D}_i \cap S | \geq m ) < \left( \frac{2^m ~ 2^{m + 1}}{n - 2(8^m)} \right)^m,
\]
and union-bound the probability of $\mathcal{A}$ ``hitting'' the secret set after $\alpha$ inferences, denoted as $p_{\text{hit}}$, with
\[
p_{\text{hit}} < \frac{\alpha}{\left( \frac{n - 2(8^m)}{2^m ~ 2^{m + 1}} \right)^m}.
\]

Here, $n - 2(8^m) = 16^m - 2(8^m) = \Theta(16^m)$, so
\[
p_{\text{hit}} < \frac{\alpha}{\Theta \left( \left( \frac{16^m}{2(4^m)} \right)^m \right)} = \frac{\alpha}{\Theta \left( 4^{m^2} \right)}.
\]

Hence, for any $\alpha = O(n^c) = O(16^{cm}) = O(4^{2c m})$, where $c$ is a constant,
\[
p_{\text{hit}} < O \left( \frac{4^{2cm}}{4^{m^2}} \right) = O(4^{-m^2}) = O(4^{-\log^2 n}).
\]

In other words, the probability that $\mathcal{A}$ does find the secret set $S$ decreases as a function of $n$.
Conditioned on this event, we upper-bound its performance with $2^{m + 1} \, f(1)$, assuming that every query is a hit.

On the other hand, if $\mathcal{A}$ never finds $S$, we further consider the following subcases: if the algorithm queries an isolated point, no marginal probability is changed; if a clump point is queried, only the marginal probabilities of the points in the same clump are updated.
The expected performance in these two cases can be upper-bounded by pretending that the algorithm had a budget of size $2 \, T = 2^{m + 2}$, half of which is spent on querying isolated points and half on clump points.

The expected utility after $T$ queries on isolated points is $\mathbb{E} \big[ f (X) \big]$, where $X = \sum_{i = 1}^T X_i$ and $\Pr(X_i = 1) = p_{\text{isolated}}$.
We further upper-bound this expectation using Jensen's inequality:
\[
\mathbb{E} \big[ f(X) \big] 
< f \big(  \mathbb{E} [X] \big)
= f \big(T ~ p_{\text{isolated}} \big)
= f \left( 2^{m + 1} (1 - 2^{-\frac{2^m}{2m^2}}) \right).
\]

The expected utility after $T$ queries on clump points is
% \begin{align*}
% \frac{f(T)}{4^m} + \left( 1 - \frac{1}{4^m} \right) \left( \frac{f(T - 1)}{4^m - 1} + \left( 1 - \frac{1}{4^m - 1} \right) (\cdots)\right) & = \frac{\sum_{i = 1}^T f(i)}{4^m} \\\\
% & < \frac{T f(T)}{4^m} \\\\
% & = \frac{f( 2^{m + 1})}{2^{m - 1}}.
% \end{align*}
\[
\frac{T\, f(1)}{4^m} + \left( 1 - \frac{1}{4^m} \right) \left( \frac{(T - 1) \, f(1)}{4^m - 1} + \left( 1 - \frac{1}{4^m - 1} \right) (\cdots)\right)
= \frac{f(1) \, \sum_{i = 1}^T i}{4^m}
= \frac{f(1) \, T \, (T - 1)}{2 \, 4^m}
= \frac{f(1) \, (2^{m + 1} - 1)}{2^m}.
\]

Combining the two subcases, we have the expected utility in the case where $\mathcal{A}$ never hits $S$ upper-bounded by
\[
f \left( 2^{m + 1} (1 - 2^{-\frac{2^m}{2m^2}}) \right) + \frac{f(2^{m + 1})}{2^{m - 1}}.
\]

With that, the overall expected utility of $\mathcal{A}$, denoted by $E_{\mathcal{A}}$ is upper-bounded by
\[
E_{\mathcal{A}} < 2^{m + 1} \, f(1) \, p_{\text{hit}} + f \left(2^{m + 1} (1 - 2^{-\frac{2^m}{2m^2}}) \right) + \frac{f(2^{m + 1})}{2^{m - 1}}.
\]

We then consider the upper bound of the approximation ratio
\[
\frac{E_{\mathcal{A}}}{\acro{OPT}} < 2 \, p_{\text{hit}} + \frac{f \left(2^{m + 1} (1 - 2^{-\frac{2^m}{2m^2}}) \right) }{2^m \, f(1)} + \frac{(2^{m + 1} - 1)}{4^m}.
\]

The first and third terms are arbitrarily small with increasing $m$.
As for the second term, L'Hôpital's rule shows that $1 - 2^{-\frac{2^m}{2m^2}} = \Theta \left( \frac{2m^2}{2^m} \right)$, so this term scales like $\Theta \left( \frac{f(4m^2)}{2^m} \right)$, which is $O \left( \frac{4m^2}{2^m} \right)$ and also arbitrarily small with increasing $m$.
As a result, algorithm $\mathcal{A}$ cannot approximate the optimal policy by a constant factor.

\section{Implementation and Pruning}
\label{sec:pruning}

As stated in \cref{subsec:model}, we use a $k$-\acro{NN} model as the probabilistic classifier in our experiments, which achieves reasonable generalization error in practice and computationally efficient.
Another benefit of this model is that it is possible to cheaply compute a \emph{posterior probability upper bound} $p^*$, given any data set $\mathcal{D}$ and an additional observation with label $y$:
\[
\max_{x' \in \mathcal{X} \setminus \mathcal{D}} p_c \big( x' \mid \mathcal{D} \cup \{ (x, y) \} \big) \leq p^*_c \left(y, \mathcal{D} \right).
\]
This upper bound is useful in that we may then bound the approximate expected terminal utility $\overline{u}$ conditioned on label $c$ when computing the score in \cref{eq:score}, i.e., $\alpha \left( x \mid y = c \right)$, and therefore the overall score $\alpha (x)$.
With these score upper bounds in hand, we employ branch-and-bound pruning strategies used in previous work \citep{jiang2018efficient,nguyen2021nonmyopic}.
Specifically, before evaluating the score $\alpha$ of a given candidate $x$, we compare the upper bound of $\alpha (x)$ against the current best score $\alpha^*$ we have found.
If $\alpha^*$ exceeds this bound, computing $\alpha (x)$ is unnecessary and we proceed to the next candidate.
Otherwise, since we have access to the \emph{conditional} score upper bounds for $\alpha \left( x \mid y = c \right)$, as we marginalize over each label $y$, we further check whether the conditional scores computed thus far, when combined with the complementary conditional upper bounds, are less than $\alpha^*$.
If this is the case, we terminate the current $\alpha$ computation ``on the fly.''
The upper bounds $p_c^*$ are cheap to evaluate, and these checks add a trivial overhead to the entire procedure in the pessimistic situation of no pruning -- in practice, this cost is well worth it.
Finally, a lazy-evaluation strategy is used, where candidates are evaluated in descending order of their score upper bounds, so that a given point that may be pruned will not be evaluated.

We also employ a pruning strategy for the inner batch-building procedure.
We first note that, in this procedure, points having the same marginal probabilities $p_c$ (conditioned on a putative query) are interchangeable as we have assumed conditional independence.
We point out one particular set of such points with equal $p_c$: those whose marginal probabilities have not been updated from the prior due to having no labeled nearest neighbors.
In a typical \acro{AS} iteration, there may be many such points, especially in early stages of the search.
Pruning duplicates appropriately allows the follow-on batch to be built more efficiently, and we empirically observed a drastic improvement in our experiments.
A welcomed property of this method is that it has the most impact in the early iterations of a search, which would usually be the longest-running iterations otherwise.
Overall, the combination of these strategies allows our algorithm to scale to large data sets (\textgreater 100\,000 points).

\section{Data Sets}
\label{sec:data}
We now describe the data sets used in our experiments in \cref{sec:experiments}.
These data sets are curated from authors of respective publications, as detailed below.
No identifiable information or offensive content is included in the data.

\begin{table}[t]
\centering
\caption{
Class-specific precision-at-$k$ statistics of the $k$-\acro{NN}, averaged across 10 randomly sampled training sets of size 499 (1\% of the data set)
}
\vskip 10pt
\begin{tabular}{cccccc}
\toprule
 &  & precision at 1 & precision at 5 & precision at 10 & precision at 50 \\
\midrule
\multirow{3}{*}{easy} & \texttt{trouser} & 0.60 & 0.60 & 0.60 & 0.59 \\
 & \texttt{bag} & 0.70 & 0.80 & 0.81 & 0.85 \\
 & \texttt{ankle boot} & 0.40 & 0.52 & 0.60 & 0.49 \\
\midrule
\multirow{3}{*}{hard} & \texttt{t-shirt} & 0.10 & 0.24 & 0.25 & 0.20 \\
 & \texttt{sandal} & 0.50 & 0.60 & 0.56 & 0.49 \\
 & \texttt{sneaker} & 0.70 & 0.68 & 0.68 & 0.66 \\
\bottomrule
\end{tabular}
\label{tab:precision}
\end{table}

\textbf{Product recommendation.}
We use the \acro{F}ashion-\acro{MNIST} data set \citep{xiao2017fashion}, which 70\,000 contains images of ten classes of fashion articles (e.g., t-shirts, trousers, pullovers, etc.\ -- 10\,000 instances for each class) to make a product recommendation problem for clothing items.
To simulate two different users (corresponding to two search problems), we select three from the ten classes to be the positive classes:
\begin{itemize}
\item User 1: \texttt{trouser}, \texttt{bag}, \texttt{ankle boot}.
\item User 2: \texttt{t-shirt}, \texttt{sandal}, \texttt{sneaker}.
\end{itemize}
For each user, we sub-sample these positive classes uniformly at random, making the positives harder to uncover; the hit rate that a random policy selects a positive point is roughly 2\%.
To build the similarity matrix used by the $k$-\acro{NN} model, we used \acro{UMAP} \citep{mcinnes2018umap} to obtain a five-dimensional embedding of the images and compute the top 100 nearest neighbors using Euclidean distance (similarity is calculated as $\exp \left( -d^2 \right)$).

While the $k$-\acro{NN} model has been found to perform well on the remaining data sets in previous studies \citep{garnett2012bayesian,jiang2017efficient,mukadum2021efficient}, with this data, we can set select as targets classes that are either easy or difficult to identify with the $k$-\acro{NN}, to simulate different levels of difficulty of search and study the effect of the quality of the $k$-\acro{NN} on the performance of our methods.
To this end, we measure the predictive performance of the $k$-\acro{NN} and report the results in \cref{tab:precision}.
We then classify the two search problems into an ``easy'' one, where the model scores highly on precision-at-$k$ metrics (particularly important in \acro{AS}), and a ``hard'' one, where the $k$-\acro{NN} scores lower.
The $k$-\acro{NN} work as well on User 2 as on User 1, specifically on instances of \texttt{t-shirt}.
This is because this class is not well-separated from the negative points. (Among the negative points, there are similar-looking classes such as pullover and shirt.)

\textbf{Photoswitch discovery.}
We consider another \acro{AS} problem introduced by \citet{mukadum2021efficient}. 
The task is to search for molecular photoswitches with two specific desirable properties: high light absorbance and long half-lives.
The collected data set contains 2049 molecules, 733 of which are found to be targets according to the criteria defined in the study.
Further, the authors partition the points into 29 groups by their respective substructures, thus defining a multiclass \acro{AS} problem with $C = 30$ (a negative class and 29 positive classes).
The number of targets in a class ranges from 0 to 121.
Following \citet{mukadum2021efficient}, we used the Morgan fingerprints \citep{rogers2010extended} of the molecules to compute the Tanimoto similarity coefficient \citep{willett1998chemical} between each pair of molecules.
These coefficients were then used to compute the nearest neighbor set of each data point.

\textbf{The $\text{CiteSeer}^\text{x}$ data set.}
We use the $\text{CiteSeer}^\text{x}$ citation network data, introduced by \citet{garnett2012bayesian}.
This data set contains 39\,788 papers published at the 50 most popular computer science conferences and journals, and the label of each paper is its publication venue.
Following \citet{fouss2007random}, we compute the ``graph principal component analysis'' and use the first 20 components to form the feature vector of each paper.
Our objective is to search for papers in machine learning and artificial intelligence proceedings.
We conduct two sets of experiments with different numbers of classes $C$.
For $C = 5$, we select papers from \acro{N}eur\acro{IPS}, \acro{ICML}, \acro{UAI}, and \acro{JMLR} as our four target classes, adding up to 5575 targets (roughly 14\% hit rate; an average of 3.5\% per class).
For $C = 10$, we additionally include papers from \acro{IJCAI}, \acro{AAAI}, \acro{JAIR}, Artificial Intelligence, and Machine Learning as targets; this totals 12\,382 targets, yielding a 31\% overall hit rate and a 3.5\% per class.

\textbf{Drug discovery.}
We experiment with a drug discovery task using a massive chemoinformatic data set.
The goal is to identify chemical compounds that exhibit selective binding activity given a protein.
The data set consists of 120 such activity classes from BindingDB \citep{liu2007bindingdb}.
For each class, there are a small number of compounds with significant binding activity -- these are our search targets.
We use the Morgan fingerprints \citep{rogers2010extended} as the feature vectors and the Tanimoto coefficient \citep{willett1998chemical} as the measure of similarity.
In each  experiment for a given value $C \in \{ 5, 10, 15 \}$, we select $(C - 1)$ out of the 120 classes uniformly at random without replacement to form the target classes.
They are then combined with 100\,000 points sampled from the ``drug-like'' entries in the \acro{ZINC} database \citep{sterling2015zinc}, which serve as the negative set, to make up our search space.
Features for these points are binary vectors encoding chemical properties, typically referred to as fingerprints.
We use the Morgan fingerprint \citep{rogers2010extended}, which has shown good performance in past studies.
The average prevalence of a target is 0.2\%; the total hit rates are thus 1\%, 2\%, and 3\% for $C = 5$, $10$, and $15$, respectively.

\begin{table}[t]
\centering
\caption{
Values for the pseudocount $\gamma_c$, as described in \cref{subsec:model}, used in the experiments in \cref{sec:experiments}.
}
\vskip 10pt
\begin{tabular}{ccccc}
\toprule
 & \acro{F}ashion-\acro{MNIST} & Photoswitch & CiteSeer$^\text{x}$ & Drug discovery \\
\midrule
$\gamma_c$ & 0.01 & 0.005 & 0.05 & 0.001 \\
\bottomrule
\end{tabular}
\label{tab:gamma}
\end{table}

We finally report the values for the pseudocount $\gamma_c$ of the $k$-\acro{NN}, as described in \cref{subsec:model}, in \cref{tab:gamma}, which roughly estimate the prevalence of each target class in each data set.

\section{Batch Utility Approximation Quality}
\label{sec:quality}

\begin{table*}[t]
\centering
\caption{
Quality and time of our approximation method against \acro{MC} sampling, averaged across 10 random repeats of $\text{CiteSeer}^{\text{x}}$ $C = 5$ experiments.
Under each setting, the approximation with the lowest error (with respect to the chosen ground truth) is highlighted \textbf{bold}, and so is the fastest method.
}
\vskip 10pt
\begin{tabular}{ccrcccrrrrr}
\toprule
 & \multicolumn{5}{c}{\acro{RMSE}} & \multicolumn{5}{c}{Time in seconds} \\
\cmidrule(lr){2-6} \cmidrule(lr){7-11}
Ground truth & \multicolumn{2}{c}{Exact} & \multicolumn{3}{c}{\acro{MC}$\left( 2^{15} \right)$} & \multicolumn{2}{c}{Exact} & \multicolumn{3}{c}{\acro{MC}$\left( 2^{15} \right)$} \\
\cmidrule(lr){2-3} \cmidrule(lr){4-6} \cmidrule(lr){7-8} \cmidrule(lr){9-11}
$b$ & 3 & \multicolumn{1}{c}{10} & 30 & 100 & 300 & \multicolumn{1}{c}{3} & \multicolumn{1}{c}{10} & \multicolumn{1}{c}{30} & \multicolumn{1}{c}{100} & \multicolumn{1}{c}{300} \\
\midrule
Exact & - & \multicolumn{1}{c}{-} & \acro{NA} & \acro{NA} & \acro{NA} & 0.0031 & 53.7149 & \multicolumn{1}{c}{\acro{NA}} & \multicolumn{1}{c}{\acro{NA}} & \multicolumn{1}{c}{\acro{NA}} \\
\acro{MC}$\left( 2^5 \right)$ & \multicolumn{1}{r}{0.0021} & 0.0027 & \multicolumn{1}{r}{0.0060} & \multicolumn{1}{r}{0.0090} & \multicolumn{1}{r}{0.0167} & 0.0026 & 0.0015 & 0.0021 & 0.0056 & 0.0195 \\
\acro{MC}$\left( 2^{10} \right)$ & \multicolumn{1}{r}{0.0004} & 0.0008 & \multicolumn{1}{r}{0.0014} & \multicolumn{1}{r}{\textbf{0.0021}} & \multicolumn{1}{r}{\textbf{0.0025}} & 0.0190 & 0.0304 & 0.0576 & 0.1574 & 0.4581 \\
\acro{MC}$\left( 2^{15} \right)$ & \multicolumn{1}{r}{\textbf{0.0001}} & \textbf{0.0001} & - & - & - & 0.4908 & 0.7933 & 1.7341 & 4.7440 & 14.4539 \\
Ours & \multicolumn{1}{r}{0.0001} & 0.0003 & \multicolumn{1}{r}{\textbf{0.0010}} & \multicolumn{1}{r}{0.0028} & \multicolumn{1}{r}{0.0059} & \textbf{0.0001} & \textbf{0.0003} & \textbf{0.0003} & \textbf{0.0003} & \textbf{0.0004} \\
\bottomrule
\end{tabular}
\label{tab:approximation5}

\bigskip

\caption{
Quality and time of our approximation method against \acro{MC} sampling, averaged across 10 random repeats of $\text{CiteSeer}^{\text{x}}$ $C = 10$ experiments.
Under each setting, the approximation with the lowest error (with respect to the chosen ground truth) is highlighted \textbf{bold}, and so is the fastest method.
}
\vskip 10pt
\begin{tabular}{ccrcccrrrrr}
\toprule
 & \multicolumn{5}{c}{\acro{RMSE}} & \multicolumn{5}{c}{Time in seconds} \\
\cmidrule(lr){2-6} \cmidrule(lr){7-11}
Ground truth & \multicolumn{1}{c}{Exact} & \multicolumn{4}{c}{\acro{MC}$\left( 2^{15} \right)$} & \multicolumn{1}{c}{Exact} & \multicolumn{4}{c}{\acro{MC}$\left( 2^{15} \right)$} \\
\cmidrule(lr){2-2} \cmidrule(lr){3-6} \cmidrule(lr){7-7} \cmidrule(lr){8-11}
$b$ & 3 & \multicolumn{1}{c}{10} & 30 & 100 & 300 & \multicolumn{1}{c}{3} & \multicolumn{1}{c}{10} & \multicolumn{1}{c}{30} & \multicolumn{1}{c}{100} & \multicolumn{1}{c}{300} \\
\midrule
Exact & - & \multicolumn{1}{c}{\acro{NA}} & \acro{NA} & \acro{NA} & \acro{NA} & 0.0056 & \multicolumn{1}{c}{\acro{NA}} & \multicolumn{1}{c}{\acro{NA}} & \multicolumn{1}{c}{\acro{NA}} & \multicolumn{1}{c}{\acro{NA}} \\
\acro{MC}$\left( 2^5 \right)$ & \multicolumn{1}{r}{0.0023} & 0.0056 & \multicolumn{1}{r}{0.0091} & \multicolumn{1}{r}{0.0141} & \multicolumn{1}{r}{0.0275} & 0.0008 & 0.0016 & 0.0022 & 0.0060 & 0.0178 \\
\acro{MC}$\left( 2^{10} \right)$ & \multicolumn{1}{r}{0.0005} & 0.0013 & \multicolumn{1}{r}{\textbf{0.0010}} & \multicolumn{1}{r}{\textbf{0.0031}} & \multicolumn{1}{r}{\textbf{0.0032}} & 0.0167 & 0.0281 & 0.0536 & 0.1522 & 0.4390 \\
\acro{MC}$\left( 2^{15} \right)$ & \multicolumn{1}{r}{\textbf{0.0001}} & \multicolumn{1}{c}{-} & - & - & - & 0.5122 & 0.8178 & 1.6678 & 4.7923 & 13.9236 \\
Ours & \multicolumn{1}{r}{0.0002} & \textbf{0.0007} & \multicolumn{1}{r}{0.0017} & \multicolumn{1}{r}{0.0055} & \multicolumn{1}{r}{0.0120} & \textbf{0.0003} & \textbf{0.0005} & \textbf{0.0004} & \textbf{0.0005} & \textbf{0.0007} \\
\bottomrule
\end{tabular}
\label{tab:approximation10}
\end{table*}

We run simulations to compare the performance of the Jensen's upper bound $\overline{u}$ against Monte Carlo (\acro{MC}) sampling, under the logarithmic utility function $u \left( \mathcal{D} \right) = \sum_{c > 1} \log \left( 1 + m_c \right)$.
Using the $\text{CiteSeer}^\text{x}$ data set described in \cref{sec:data}, we first construct a training data set $\mathcal{D}$ by randomly selecting $50$ points for each class ($\vert \mathcal{D} \vert = 50 \, C$), and compute the posterior probabilities $p_c$ with the $k$-\acro{NN} to simulate a typical \acro{AS} iteration.
A random target batch $X$ of size $b$ is then chosen, and the considered approximation methods are applied on this batch.
This entire procedure is repeated 10 times for each setting of $(C, b)$, and the average root mean squared error (\acro{RMSE}) and time taken for each method to return are reported in \cref{tab:approximation5,tab:approximation10}.
($\acro{MC}(s)$ denotes \acro{MC} sampling using $s$ samples.)
Note that in settings with large $C$ or $b$, exact computation of \cref{eq:exact} is prohibitively expensive, in which case $\acro{MC} \left( 2^{15} \right)$ is used as the ground truth.
Overall, our Jensen's approximation offers a good tradeoff between accuracy and speed.

\section{Experimental Details \& Further Results}
\label{sec:more_results}

\begin{table}[t]
\centering
\caption{
Square root utility and standard errors across 20 repeated experiments for each setting.
$C$ is the total number of unique classes in the search space.
The best performance in each column is highlighted in \textbf{bold}; policies not significantly worse than the best (according to a two-sided paired $t$-test with a significance level of $\alpha = 0.05$) are in \emph{\color{blue} blue italics}.
}
\vskip 10pt
\begin{tabular}{cccc}
\toprule
 & CiteSeer$^\text{x}$ & \multicolumn{2}{c}{Drug discovery} \\
\cmidrule(lr){2-2} \cmidrule(lr){3-4}
 & $C = 10$ & $C = 10$ & $C = 15$ \\
\midrule
\citet{he2007nearest} & \resultalt{35.51}{0.22} & \resultalt{9.40}{0.33} & \resultalt{13.60}{0.45} \\
\citet{vanchinathan2015discovering} & \resultalt{48.61}{1.22} & \resultalt{32.28}{1.26} & \resultalt{41.10}{1.99} \\
\citet{malkomes2021beyond} & \resultalt{31.32}{0.01} & \resultalt{18.26}{0.79} & \resultalt{25.82}{0.87} \\
\midrule
\acro{ENS} & \resultalt{59.31}{0.81} & \resultalt{32.12}{1.17} & \resultalt{36.71}{1.71} \\
\acro{RR-greedy} & \resultalt{57.37}{0.33} & \resultalt{30.66}{1.82} & \resultalt{37.51}{1.38} \\
\acro{RR-UCB} & \begin{tabular}[c]{@{}c@{}}\resultalt{57.87}{0.32}\\ ($\beta^* = 3$)\end{tabular} & \begin{tabular}[c]{@{}c@{}}\resultalt{30.77}{1.47}\\ ($\beta^* = 1$)\end{tabular} & \begin{tabular}[c]{@{}c@{}}\resultalt{38.02}{1.49}\\ ($\beta^* = 3$)\end{tabular} \\
\acro{RR-ENS} & \resultalt{60.20}{0.30} & \resultblueitalt{37.09}{1.20} & \resultalt{45.05}{1.27} \\
\midrule
One-step & \resultalt{60.59}{0.46} & \resultblueitalt{38.86}{1.59} & \resultalt{46.60}{1.70} \\
\acro{DAS} (ours) & \resultboldalt{62.41}{0.24} & \resultboldalt{38.89}{1.76} & \resultboldalt{55.50}{1.41} \\
\bottomrule
\end{tabular}
\label{tab:sqrt_results}
\end{table}

We first detail how we set the hyperparameters of the active learning baselines used in our experiments.
\begin{itemize}
\item The policy by \citet{vanchinathan2015discovering} has two hyperparameters: $\lambda$ encourages diversity (measured by the logdet of the Gram matrix of the collected data), and $\beta_t$ encourages \acro{UCB}-style exploration of the space.
We run all variants of this policy with $\lambda \in \{ 0.25, 0.5, 0.75 \}$ and $\beta_t \in \{ 0.1, 0.3, 1, 3, 10 \}$ and report the best performance in \cref{tab:results,tab:sqrt_results}.
\item The policy by \citet{he2007nearest} has a hyperparameter $p$, which is an estimate of the prevalence of the targets within the search space.
We set this value in the same way as we set $\gamma_c$ in \cref{tab:gamma}.
\item The policy by \citet{malkomes2021beyond} has a hyperparameter $r$, which sets the radius of the spheres that compute their coverage measure.
As we have access to similarity scores (between $0$ and $1$) among points in each of our data sets, the spheres in this method cover points that are sufficiently similar (similarity greater than threshold $1 - r$) to a collected target.
We run all variants of this policy with $r \in \{ 0.25, 0.5, 0.75 \}$ and report the best performance in \cref{tab:results,tab:sqrt_results}.
\end{itemize}

We also include our experiment results with the square root utility function
\[
u \left( \mathcal{D} \right) = \sum_{c > 1} \sqrt{m_c},
\]
under the CiteSeer$^\text{x}$ $C = 10$ and drug discovery $C \in \{ 10, 15 \}$ settings in \cref{tab:sqrt_results}.
Again, our method \acro{DAS} was applied to this utility function without any algorithmic modification from the logarithmic utility.
Under this utility, \acro{DAS} still consistently performs the best across the investigated settings, showing that our proposed method generalizes well to this utility.

\begin{table}
\centering
\caption{Logarithmic utility and standard errors across 20 repeated experiments on the Fashion \acro{MNIST} data.}
\label{tab:similar}
\vskip -0.1cm
\begin{tabular}{ccc}\\
\toprule  
 & easy & hard \\
 \midrule
\acro{DAS} & 14.02 (0.06) & 12.38 (0.30) \\
\acro{SIMILAR} & 13.89 (0.02) & 12.31 (0.58) \\
\bottomrule
\end{tabular}
\end{table}

Finally, we present brief results comparing the \acro{SIMILAR} algorithm \citep{kothawade2021similar} (instantiated for the task of rare class detection) with our method \acro{DAS}.
A neural network-based policy, \acro{SIMILAR} can be readily applied to our production recommendation problem with the Fashion \acro{MNIST} data, and the results are presented in \cref{tab:similar}, where we note a slight degradation in performance going from \acro{DAS} to \acro{SIMILAR}.

\section{Extension to Within-Class Diversity-Aware Active Search}
\label{sec:within}

In many important applications, one is concerned about not only diversity in the observed labels, but also a measure of within-class diversity, defined in the feature space.
In this section, we describe how our diversity-aware active search framework can be extended to this setting.
First, we use a monotone submodular set function $g$ (e.g., logdet) to quantify the diversity of a set of class-$c$ positives $X_c$.
Then, $g \left( X_c \right)$, instead of the class count $m_c$, is used as input to the class-specific concave utility $f_c$ -- our final utility is $\sum f_c \left( g(X_c) \right)$.
This way, when a new point is added to $X_c$, the more different the point is from $X_c$, the more $g$, and in turn $f_c$, increases; this modified utility thus captures within- and cross-class diversity.
Conveniently, our method still applies: $\mathbb{E} \big[ f_c \left( g( X_c) \right) \big] < f_c \big( \mathbb{E} \left[ g(X_c) \right] \big)$ (Jensen's inequality), and our lookahead seeks a batch maximizing $\overline{u} = \sum f_c \big( \mathbb{E} \left[ g(X_c) \right] \big)$.
The expectation of the submodular $g(X_c)$ is itself submodular, and thus the sum $\overline{u}$ is submodular and can be greedily optimized.
In other words, our nonmyopic decision-making scheme remains viable.
(In the special case of the count function $g(X_c) = | X_c |$, this reduces to what the main text presents.)

\section{Software}
\label{sec:license}

Matlab implementation of this work is released under \acro{MIT} license at \url{https://github.com/KrisNguyen135/diverse_as}.

\end{document}